\documentclass[10pt,twocolumn,letterpaper]{article}

\usepackage{iccv}              % To produce the CAMERA-READY version
\usepackage{makecell}

\usepackage{multirow}
\usepackage{subcaption}
\usepackage{placeins} % Add this in your preamble

\usepackage[dvipsnames]{xcolor}

\usepackage{booktabs}
\usepackage{tikz}

\usepackage{bbm}
\definecolor{cvprblue}{rgb}{0.21,0.49,0.74}
\usepackage[pagebackref,breaklinks,colorlinks,citecolor=cvprblue]{hyperref}
\usepackage{epigraph} 
\usepackage{placeins} 

\usepackage{listings}

\usepackage[para]{footmisc}

\def\paperID{14325} % *** Enter the Paper ID here
\def\confName{ICCV}
\def\confYear{2025}

\title{ImageGem: In-the-wild Generative Image Interaction Dataset for Generative Model Personalization}

\author{Yuanhe Guo$^{1*}$, Linxi Xie$^{1*}$, Zhuoran Chen$^{1}$, Kangrui Yu$^{1}$, \\
Ryan Po$^{2}$, Guandao Yang$^{2}$, Gordon Wetztein$^{2}$, Hongyi Wen$^{1\dag}$\\
{$^{1}$NYU \hspace{1em} $^{2}$Stanford}\\
{\url{https://maps-research.github.io/imagegem-iccv2025/}}
}

\begin{document}

\twocolumn[{
\maketitle
\centering
\vspace*{-0.73cm}
\includegraphics[width=0.95\textwidth]{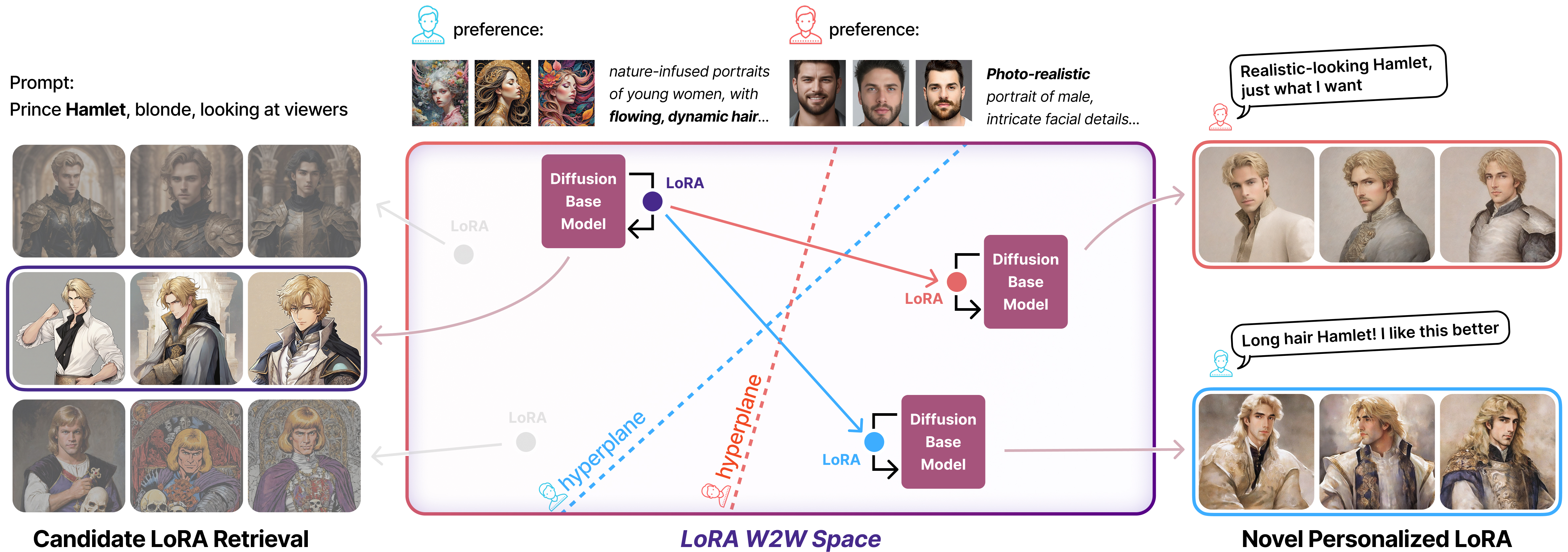}
\captionof{figure}{Our proposed ImageGem dataset and its applications. The left side illustrates image and generative model retrieval. On the right, we demonstrate a novel task of generative model personalization through LoRA weights-to-weights (W2W) space construction.}
\label{fig:teaser}
\vspace*{0.27cm}
}]

\def\thefootnote{*}\footnotetext{Equal contribution.}
\def\thefootnote{\dag}\footnotetext{Correspondence.}
\def\thefootnote{\arabic{footnote}}

\begin{abstract}
We introduce ImageGem, a dataset for studying generative models that understand fine-grained individual preferences.
We posit that a key challenge hindering the development of such a generative model is the lack of in-the-wild and fine-grained user preference annotations. 
Our dataset features real-world interaction data from 57K users, who collectively have built 242K customized LoRAs, written 3M  text prompts, and created 5M generated images. 
With user preference annotations from our dataset, we were able to train better preference alignment models. 
In addition, leveraging individual user preference, we investigated the performance of retrieval models and a vision-language model on personalized image retrieval and generative model recommendation. 
Finally, we propose an end-to-end framework for editing customized diffusion models in a latent weight space to align with individual user preferences.
Our results demonstrate that the ImageGem dataset enables, for the first time, a new paradigm for generative model personalization.

\end{abstract}   
\section{Introduction}
\label{sec:intro}

\epigraph{A thousand Hamlets in a thousand people's eyes}{}

Recent advances in text-to-image models~\cite{rombach2022high} have empowered users to generate a portrait of Hamlet, Shakespeare's famous character, from merely a short and under-specified description like ``A portrait of Hamlet''.
However, each user has their imagination about the portrait of Prince Hamlet.
Can generative models capture and produce the version that aligns with individual preference?

Current progress toward building personalized text-conditioned generative models is mainly driven by data availability.
For example, with the presence of datasets that contain different person or object identities, prior works are able to customize diffusion model to generate these user-specified concepts.~\cite{ruiz2023dreambooth,ruiz2024hyperdreambooth, qian2024omni, wu2025fiva}. 
These works, however, do not address under-specified inputs that require reasoning about individual preference, such as generating an image of ``my favorite dog.''
Similarly, enabled by datasets with user preference annotations~\cite{wang2022diffusiondb, kirstain2023pick, liang2024rich, wu2023human}, many works are able to create text-to-images models that align with human preference~\cite{wallace2024diffusion, xu2023imagereward, fan2023dpok}.    
These methods, however, focus on aggregated preference, such as generating an image that the general population will favor.  
How to create a generative model aligned with personal preferences remains under-explored due to the lack of large-scale and fine-grained user preference annotations.
Existing efforts toward this end are thus limited to zero-shot approaches~\cite{salehi2024viper}, which usually require user input during inference time.
Such zero-shot approaches find it difficult to leverage similarity among users.
As a result, they can be expensive and limited to a few predetermined dimensions of individual preference.

Motivated by the gap between aggregated preference modeling and personalization at the individual level, we propose \textit{ImageGem} dataset, the first large-scale dataset that contains diverse user behaviors from real-world users employing their generative models.
We sourced our data from Civitai~\footnote{https://civitai.com/}, one of the most popular AIGC platforms where users create and publicly share both their customized models and generated images.
In addition to the content filter provided by Civitai, we further evaluated the safety of images and prompts, labeling them accordingly to ensure a reliable dataset for downstream tasks. 

We setup a few evaluations on the quality of aggregated and individual preference data from our dataset. 
Specifically, we train SD1.5~\cite{rombach2022high} with DiffusionDPO~\cite{wallace2024diffusion} on aggregated preference data and demonstrate improved image quality over a widely-used dataset for preference alignment~\cite{kirstain2023pick}.
We further leverage individual preference data to examine the quality of personalized image retrieval and generative model recommendations with retrieval-based models. 
In addition to retrieval-based models, we leverage a vision-language model (VLM)~\cite{agrawal2024pixtral12b} for user preference captioning and ranking by prompting the VLM to generate structured descriptions.
We demonstrate that integrating VLM enhances ranking interpretability.

Our dataset enables a new application of \textit{Generative Model Personalization}, where customized diffusion models (e.g. LoRA~\cite{hu2022lora}) are created to align with individual preference. 
We leverage a subset of user-created LoRAs from ImageGem to construct a latent weight space. 
By capturing individual preferences from historical user-generated images, we learn editing directions in this latent space, enabling progressive model adaptation to user preferences.
We summarize our contributions as the following:
\begin{itemize}
    \item We present the first large-scale dataset consisting of user fine-grained preferences towards generative models and images. Our dataset consists of metadata such as prompts, images, and user feedback. We apply safety checks on metadata and ensure the diversity of our dataset.
    \item We evaluate the quality of our curated dataset and showcase its preference labels for several downstream applications, including general preference alignment, personalized image retrieval, and generative recommendation.
    \item We propose an end-to-end framework for editing customized diffusion models toward individual user preference, demonstrating a new application in curating personalized generative models.
\end{itemize}

\section{Related Work}

\subsection{Dataset for Preference Alignment}
Several works investigate how to train better diffusion models that align with human preference~\cite{kirstain2023pick, liang2024rich, wu2023human}. For example, Pick-a-Pic~\cite{kirstain2023pick} collected ratings on image pairs from about 6K users and demonstrated their PickScore achieved state-of-the-art alignment with human judgments on image generations. 
RichHF-18K~\cite{liang2024rich} exemplifies the heterogeneous user feedback such as predicted scores and heatmaps can be leveraged to improve RLHF. 
FiVA~\cite{wu2025fiva} curated a fine-grained visual attributes dataset of 1 million generated images with detailed annotations.

Our dataset connects to this line of research on developing human preference datasets to improve image generation models, but with several key differences. Our dataset contains observational data from a cohort of in-the-wild users, e.g., interaction logs between 57K users and 242K generative models from 2023/09 to 2025/01. Generative models in our dataset are up-to-date and customized by users, reflecting real-world usage and preferences. Moreover, our dataset captures individual-level user preference as opposed to aggregated-level preference.

\subsection{Personalizing Generative Models}
A group of work personalizes generative models through capturing prompt instructions~\cite{Brooks2022InstructPix2PixLT} or other diverse modality of user inputs~\cite{Zhang2023ControlNet}. Another branch widely used approaches enable efficient fine-tuning given personal preference, starting with textual inversion~\cite{gal2023TextualInversion} and DreamBooth~\cite{ruiz2023dreambooth}. Further works train adapters for preserving specific identities such as human faces~\cite{wang2024instantid, zhang2024flashface}.
Weights2Weights~\cite{dravid2024interpreting} demonstrates the feasibility of customizing models in a latent weight space towards certain attributes by creating a dataset of 60k LoRAs from human faces.
ViPer~\cite{salehi2024viper} proposes a zero-shot user preference learning framework via a two-stage process: first capturing the user's general preferences through their comments, and then a vision-language model extracts structured preferences and is used for editing the prompts for text-to-image generation. 

Our dataset augments previous work by providing fine-grained user preference data and user-customized diffusion models in large scale. 
With this dataset, we provide a new perspective to personalizing visual generative models by editing pre-trained models to individual user preference captured from unspecified interaction data, which is a more challenging and realistic task.

\begin{figure*}[t!]
    \centering
    \begin{subfigure}[b]{0.49\textwidth}
        \includegraphics[width=\textwidth]{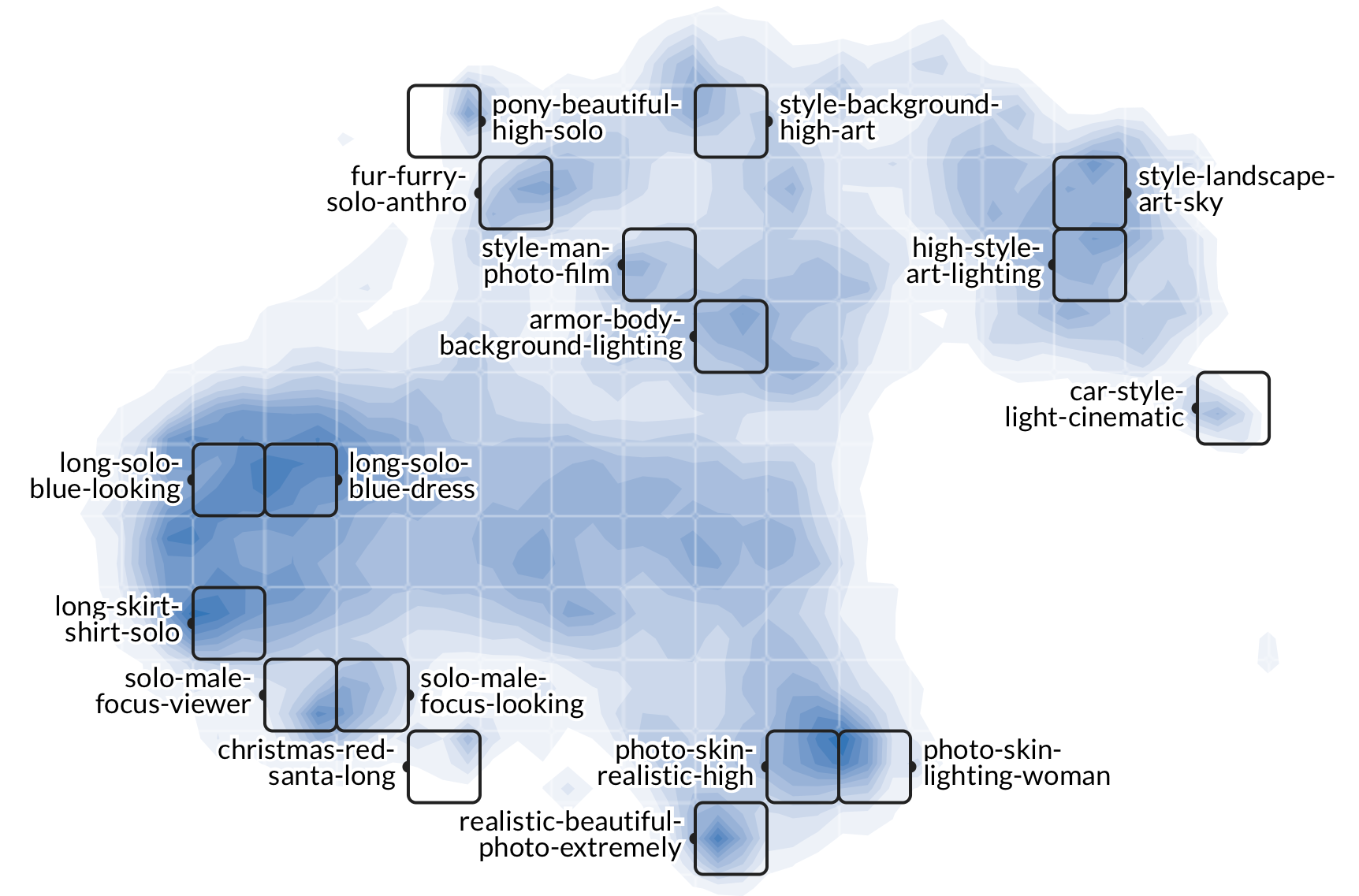}
        \caption{Contour plot of 1M images sampled from our dataset, visualized using UMAP to reduce the dimensionality of their CLIP embeddings.}
        \label{fig:wizmap_images}
    \end{subfigure}
    \hfill 
    \begin{subfigure}[b]{0.49\textwidth}
        \includegraphics[width=\textwidth]{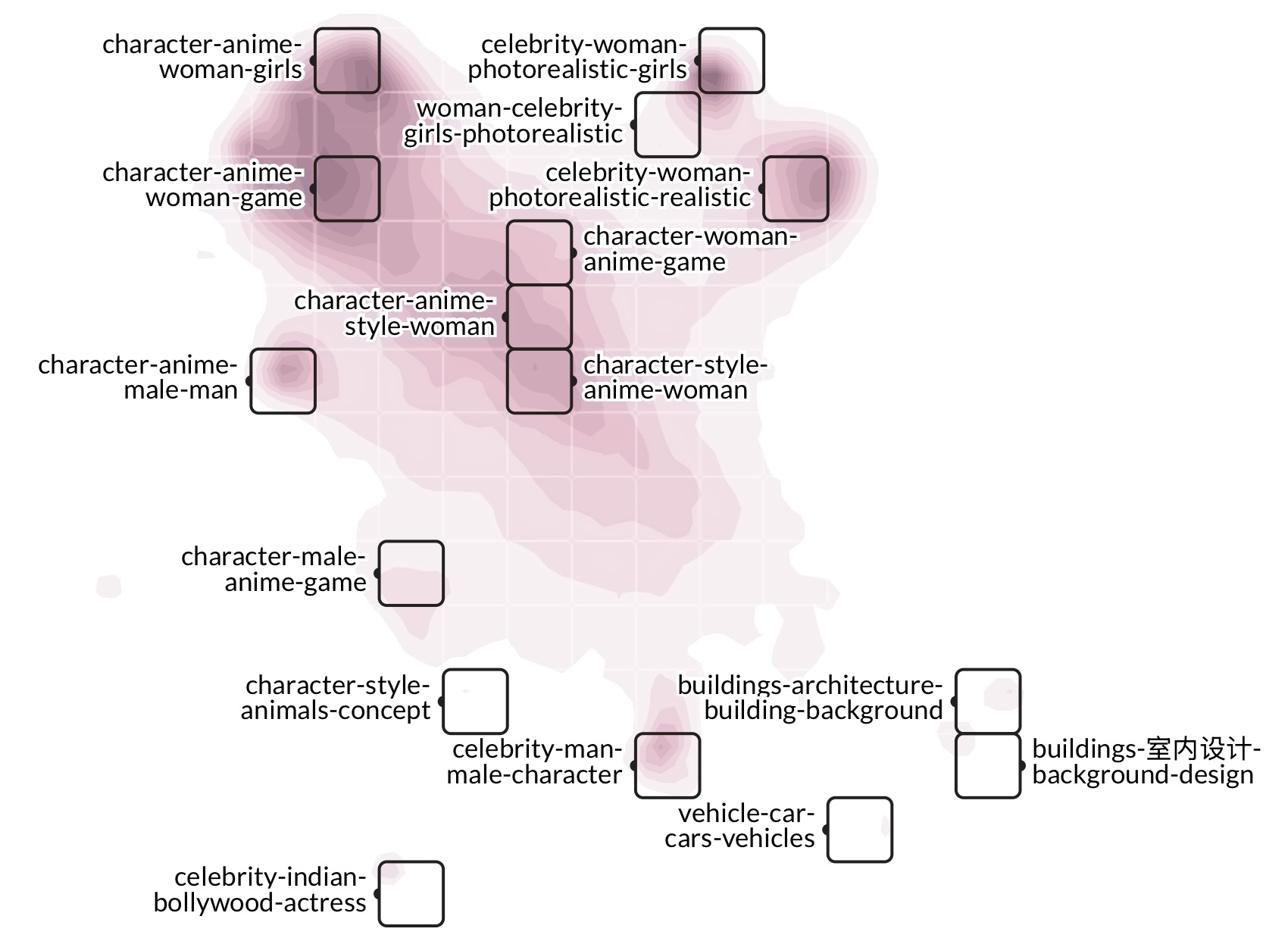}
        \caption{Contour plot of LoRA model checkpoints, where each LoRA is represented by the mean embedding of its corresponding images.}
        \label{fig:wizmap_loras}
    \end{subfigure}
    \caption{WizMap~\cite{wangWizMapScalableInteractive2023}-based visualization for our ImageGem dataset, divided into two parts. The left panel shows a UMAP embedding of 1M images sampled from the dataset, while the right panel illustrates a contour plot of LoRA model checkpoints. Both visualizations use grid tiles to display key words extracted from image prompts or model tags.}
    \label{fig:enter-label}
\end{figure*}

\section{The ImageGem Dataset}

\begin{table*}[t]
    \centering
    \setlength{\tabcolsep}{3pt}
    \renewcommand{\arraystretch}{0.8}
    \resizebox{\textwidth}{!}{
        \begin{tabular}{lccccccccc}
        \toprule
         & \thead{Images} & \thead{Unique Prompts} & \thead{LoRA Model \\ Checkpoints} & \thead{Unique \\ Model Tags} & \thead{Total Users} & \thead{Model Uploaders} $*$ & \thead{Avg Images \\ Per Uploader} $\dagger$ & \thead{Avg Models \\ Per Uploader} $\dagger$ & \thead{Avg Images \\ Per Model} $\ddagger$ \\
        \midrule
        Raw & 5,658,107 & 2,975,943 & 242,889 & 105,788 & \multirow{2}{*}{57,245} & 19,003 & 49 & 12 & 62 \\
        Filtered & 4,916,134 & 2,895,364 & 242,118 & 97,434 & & 18,889 & 48 & 13 & 54 \\
        \bottomrule
        \end{tabular}
    }
    \caption{Statistics of our ImageGem dataset. $*$ While every user generated at least one image, not all users uploaded LoRAs. $\dagger$ Excluded highest-uploader counts for unbiased averages. $\ddagger$ Many-to-many image-model relationships may cause image double-counting.}
    \label{tab:safety_count}
\end{table*}

We constructed the \textbf{ImageGem dataset} by sourcing data from Civitai, an open-source platform for sharing fine-tuned model weights and images. This dataset captures real-world interactions between users and image generation models, offering a unique opportunity to study personalized preferences in diffusion-based systems. Below, we detail its construction, curation, and key characteristics.

\subsection{Metadata and Relational Database}

Civitai serves as a comprehensive source of user-generated content, featuring personalized diffusion models, images, and associated metadata. To build ImageGem, we leveraged Civitai’s public API, which provides information about licenses and NSFW (Not Suitable For Work) classifications. Prior to data collection, we obtained institutional IRB approval to ensure ethical compliance. 

Our dataset captures three core components: LoRA models (light-weight adapters for fine-tuning diffusion models), images generated using these LoRAs, and users who upload images and models.
To enable flexible querying and analysis, we established a ternary relationship between images, LoRAs, and users, allowing for efficient retrieval of user-specific preferences and model interactions.

\subsection{Safety Check}
Given the open nature of Civitai, we implemented rigorous safety checks to ensure the dataset’s reliability for downstream tasks. While Civitai categorizes images based on NSFW levels~\footnote{https://education.civitai.com/civitais-guide-to-content-levels/}, prompts and user-labeled LoRA tags lack explicit ratings. To address this, we used Detoxify~\cite{Detoxify}, a multilingual toxic text classifier, to estimate NSFW probabilities for prompts. A detailed distribution of NSFW probability in each aspect is shown in Appendix Fig.~\ref{fig:safety_prompt}. Images whose prompt's unsafe probabilities above 0.2 were excluded. These steps ensured that the final dataset balances diversity with safety, making it suitable for research on preference learning and model personalization.

\subsection{Dataset Overview}

Tab.~\ref{tab:safety_count} shows the essential numbers in our dataset before and after safety filtering. All 4,916,134 filtered images, which have associated prompts recorded in the metadata and are accessible from the Civitai website as of March, 2025. In the following paper, we focus on analyzing the safety checked dataset.

\textbf{Images.} We computed the CLIP~\cite{Radford2021CLIP} embedding for all images, and used UMAP~\cite{McInnes2018UMAP} to reduce the dimension to 2D for visualization. The distribution shown in Fig.~\ref{fig:wizmap_images} illustrates the wild variety of topics covered in our dataset.
Recent methods, such as Compel~\cite{compel}, encode long prompts within 77 tokens with CLIP~\cite{Radford2021CLIP} using prompt weighting techniques, which makes token counting less accurate. As a result, we count the number of words in each prompt, with the average word count being 48.5. Additionally, we observe some prompts with exceptionally large word counts. The distribution of prompts with word counts exceeding 200 is shown in Appendix Fig.~\ref{fig:promp_word_count}.
Image feedback are captured by various emojis, including thumbs-up, heart, laugh, and cry. The distribution of each type of user feedback is shown in Appendix Fig.~\ref{fig:image_stats}.

\textbf{LoRA Models.} In our dataset, LoRA models are fine-tuned based on $37$ different base model structures, with $41\%$ being SD 1.5~\cite{rombach2022high}, $31\%$ Pony~\footnote{AstraliteHeart/pony-diffusion}, $12\%$ SDXL 1.0~\cite{podell2024sdxl} and $9\%$ Flux.1~\cite{flux2024}. Fig.~\ref{fig:wizmap_loras} shows the distribution of LoRA models. We represent each LoRA by averaging the embeddings of its image embeddings, and the text labels are tags labeled by model uploaders.

\textbf{User Interactions.}
Our dataset includes two types of user interaction data: (1) \textit{Individual level user-model interactions}, which capture user-specific image generation configurations (e.g., prompts) to analyze individual-level preference, usage patterns, and prompting strategies across LoRA. This include $1,739,947$ in-house images created by LoRA up-loaders to showcase their model capability. The remaining $3,176,187$ images serve as historical records of user preferences, enabling tasks like image and model recommendation. (2) \textit{Aggregate level user-image feedback}, which provide aggregated emoji feedback (e.g. like/dislike) for content filtering and benchmarking preference alignment methods.

\section{Applications and Methods}

We now present three applications followed by the creation of the ImageGem dataset, focusing on the aggregate-level and individual-level user interaction data accompanied by other metadata to enable various applications.

\subsection{Aggregated Preference Alignment}

With in-the-wild user preference annotations (e.g., likes, crys) towards prompt-image pairs from our dataset, a direct application of our dataset is to train preference alignment models. Following DiffusionDPO~\cite{wallace2024diffusion}, for preference pairs $(\mathbf{c}, \mathbf{x_0^w}, \mathbf{x_0^l})$ of prompt $\mathbf{c}$, and image preference label $\mathbf{x_0^w} \succ \mathbf{x_0^l}$, the training objective is as following:

\begin{equation}
\begin{aligned}
\max_{p_{\theta}} \quad & \mathbb{E}_{c \sim \mathcal{D}_c, \mathbf{x}_{0:T} \sim p_{\theta}(\mathbf{x}_{0:T} | \mathbf{c})} 
[ r(\mathbf{c}, \mathbf{x}_0) ]  \\
& \quad - \beta \mathbb{D}_{\text{KL}} [ p_{\theta}(\mathbf{x}_{0:T} | \mathbf{c}) \| p_{\text{ref}}(\mathbf{x}_{0:T} | \mathbf{c}) ].
\end{aligned}
\end{equation}

Here, $p_{\text{ref}}$ is a pre-trained diffusion model, and $p_{\theta}$ is the updated model to align with preferences with trainable parameters $\theta$, and $T$ being the diffusion timestep. The reward function $r(\mathbf{c}, \mathbf{x_0})$ is defined as:

\begin{equation}
    \begin{aligned}
        r(\mathbf{c}, \mathbf{x_0}) = \mathbb{E}_{p_{\theta}(\mathbf{x}_{1:T}|\mathbf{x}_0,\mathbf{c})}[R(\mathbf{c}, \mathbf{x}_{0:T})],
    \end{aligned}
\end{equation}
with $R(\mathbf{c}, \mathbf{x}_{0:T})$ being the reward on the whole chain.

Different from other large-scale datasets such as Pick-a-Pic that rely on human annotations of explicit preference over image pairs, the preference pairs in our dataset are from natural observations and are implicit.
To curate high-quality preference pairs, we first cluster the prompts' CLIP embeddings in our dataset with HDBScan~\cite{McInnesLeland2017hHdb}. Within each cluster, we construct preference pairs using min-max pairing over Human Preference Score v2 ~\cite{wu2023human}. Through comparisons with Pick-a-Pic on various metrics~\cite{kirstain2023pick, wu2023human, hessel2022clipscore}, we demonstrate that our dataset achieves improved aggregate-level preference alignment in Sec.~\ref{sec:exp-dpo}.

\subsection{Retrieval and Generative Recommendation}

With abundant individual-level preference data in our dataset, we explore personalized image retrieval~\cite{datta2008image} and generative model recommendations~\cite{guo2024gemrec}. 
For both image and generative model items, we formulate the recommendation task on the large item corpus size of millions following a two-stage retrieval-ranking paradigm~\cite{covington2016deep}, where we use collaborative filtering (CF) to retrieve a subset of top-k next interacted items given user, and then use a complex visual-language model (VLM) for ranking on that subset. 

\textbf{Candidate Retrieval.}
For large-scale image retrieval, we employ FAISS~\cite{douze2025faisslibrary} for Approximate Nearest Neighbors (ANN) search, enabling efficient and scalable vector search across millions of items. To mitigate the sparsity of the training data, we initialize the ID embedding of each image by encoding each image with a pre-trained ViT~\cite{dosovitskiy2021an}. We evaluate the performance of ItemKNN~\cite{deshpande2004item}, Item2Vec~\cite{barkan2016item2vec}, and a two-tower model~\cite{yi2019sampling} for the retrieval task.

As for candidate model retrieval, we evaluate various approaches that capture diverse facets of user preferences for generative models. 
UserKNN and ItemKNN compute user/item similarity from historical interaction data using cosine similarity and aggregate similar user/item ratings on the target item.  SASRec~\cite{kang2018selfattentivesequentialrecommendation} utilizes self-attention mechanisms to model the sequential nature of user interactions, considering order and temporal dynamics. 

\textbf{Generative Recommendation.}
We explore generative recommendation by building a VLM-based recommendation workflow that generates structured user preference descriptions as representations of user interests. We selected Pixtral-12B \cite{agrawal2024pixtral12b} as our VLM due to its ability to process multiple images and texts within one prompt, making it convenient for multi-item captioning and ranking tasks. 

Our workflow consists of two stages: item captioning and ranking. In the captioning stage, given the user's historical preferences, we prompt the VLM to generate textual representations of user's visual preference profile, denoted as $q_i$. 
For image items, the VLM extracts common features from user-generated images. For model items, we select the most-liked user-provided prompt for each model, and prompt the VLM to summarize its key attributes. 

In the ranking stage, we construct a prompt that instructs the VLM to compare \( q_i \) with each item in \( C_i \) and generate a similarity score along with an explanation, where \( C_i \) denotes the CF candidate set from the retrieval stage. 
However, VLM ranking exhibits instability, as item scores may vary across different inference requests or ranking orders. To mitigate this, we designed templates with detailed ranking criteria and adopted a randomized scoring strategy under VLM input constraints. Details in VLM prompting and our scoring strategy are included in Appendix~\ref{sec:appendix-vlm-ranking}.

\subsection{Generative Model Personalization}

Given the limitations of the retrieval-based recommendation paradigm, we propose a new framework of \textit{Generative Model Personalization}, which generates personalized LoRA models aligned with individual preferences. Building on previous work that explores a LoRA Weights2Weights (W2W) space for identity editing \cite{dravid2024interpreting}, we adapt this method to model personalization. Specifically, we use a set of user-created LoRAs to construct a latent LoRA weight space and learn editing directions that reflect user preferences. These directions can then be used to transform any LoRA model within the space, producing a personalized version without requiring re-training.

To create a W2W space, we first need to reduce and standardize the set of user-created LoRAs, we apply singular value decomposition (SVD) to each LoRA weight matrix, and retain only the top-1 component. We then flatten and concatenate the reduced matrices from all layers to obtain a vector representation \( \theta_i \in \mathbb{R}^d\) for each LoRA. This yields a dataset \( D = \{\theta_1, \theta_2, \dots, \theta_N\} \), where each point represents a distinct preference of an individual. To reduce dimensionality and identify meaningful subspaces, we applied Principal Component Analysis (PCA) on the dataset, retaining the top \( m \) principal components. This process established a basis of vectors \( \{w_1, w_2, \dots, w_m\} \), where each basis vector inherently encodes user preference, ensuring that all modifications remain within the user preference space.

To generate personalized LoRAs, we sought a direction \( v \in \mathbb{R}^d \) in the weight space that captures individual preference. Using binary labels for each user-model pair (e.g. preferred/not preferred) obtained from our dataset, we trained linear classifiers with model weights as input features. The hyperplane determined by the classifier separates the models according to whether a target user likes it or not, and the normal vector \( v \) to this hyperplane serves as the traversal direction. Given a model weight \( \theta \), tuning is achieved by moving orthogonally along the direction \( v \). The edited weights are calculated as $\theta_{\text{edit}} = \theta + \alpha v$, where \( \alpha \) is a scalar controlling the strength of the tuning operation. This adjustment modifies the model to approximate individual preferences while preserving other features.
\section{Experiments}

\begin{table}[b]
    \centering
    \setlength{\tabcolsep}{3pt}
    \renewcommand{\arraystretch}{0.9}
    \resizebox{\columnwidth}{!}{
        \begin{tabular}{lllcc}
            \toprule
            \thead[l]{Dataset (Subset)} & \thead[l]{Key Words} & \thead{\#Pairs} \\
            \midrule
            Pick-a-pic Cars & ``cars'', ``car'', ``vehicle'', ``vehicles'' & 13,436 \\
            ImageGem(Ours) Cars Small & ``cars'', ``car'', ``vehicle'', ``vehicles'' & 13,436 \\
            ImageGem(Ours) Cars Large & ``cars'', ``car'', ``vehicle'', ``vehicles'' & 27,837 \\
            \midrule
            ImageGem(Ours) Dogs & ``dog'', ``dogs'', ``puppy'', ``puppies'' & 8,764 \\
            Pick-a-pic Dogs & ``dog'', ``dogs'', ``puppy'', ``puppies'' & 10,184 \\
            \midrule 
            Pick-a-pic Scenery & ``scenery'', ``landscape'' & 12,024 \\
            ImageGem(Ours) Scenery Small & ``scenery'', ``landscape'' & 9,498 \\
            ImageGem(Ours) Scenery Large & ``scenery'', ``landscape'' & 85,473 \\
            \bottomrule
        \end{tabular}
    }
    \caption{Subsets of three topics sampled from Our dataset and Pick-a-Pic at comparable scales, with images filtered by key words and excluding human character-related prompts. For cars and scenery, we found significantly more image pairs than Pick-a-Pic, so we split it into a small set and a large set for ablation study.}
    \label{tab:subsets}
\end{table}

\begin{table}[b]
    \centering
    \scriptsize
    \setlength{\tabcolsep}{3pt}
    \renewcommand{\arraystretch}{0.9}
    \resizebox{\columnwidth}{!}{
    \begin{tabular}{lrrrrrr}
        \toprule
        \textbf{Dataset (Subset)} & \textbf{\#Users} & \textbf{\#Items} & \textbf{\#Interactions} & \textbf{\#Avg.Seq} & \textbf{Sparsity} \\
        \midrule
        Images - 1M           & 15,917 & 1,002,796 & 1,002,796 & 63.00 & 99.99\%  \\
        Models - 200K  & 10,364  & 53,590 &205,160 & 19.80 & 99.96\%  \\
        \bottomrule
    \end{tabular}
    }
    \caption{Dataset overview for the image retrieval and generative recommendation tasks.}
    \label{tab:rec-dataset}
\end{table}

\subsection{Aggregated Preference Alignment}
\label{sec:exp-dpo}

\begin{figure*}[t!]
    \centering
    \includegraphics[width=\textwidth]{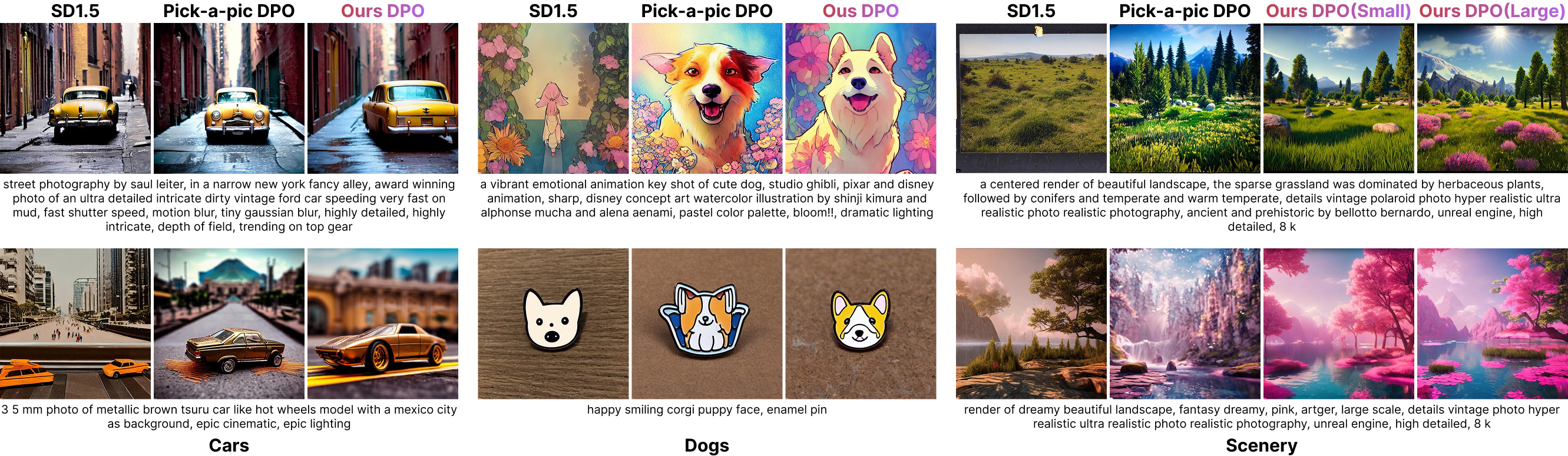}
    \caption{Qualitative DiffusionDPO results comparison of images generated with OOD prompts in three topics sampled from DiffusionDB. For each prompt, random seed and all other hyperparameters are kept the same. Zoom in for the best view.}
    \label{fig:dpo_quali}
\end{figure*}

\begin{table}[ht]
    \centering
    \scriptsize
    \setlength{\tabcolsep}{3pt}
    \renewcommand{\arraystretch}{0.9}
    \resizebox{\columnwidth}{!}{
        \begin{tabular}{llccc}
            \toprule
            \thead[l]{Dataset (Subset)} & \thead{Pick Score $\uparrow$} & \thead{HPSv2 $\uparrow$} & \thead{CLIP Score $\uparrow$} \\
            \midrule
            Original SD1.5 & 0.1977 & 0.2637 & 0.3581 \\
            Pick-a-pic Cars & 0.1993 & 0.2690 & 0.3607 \\
            ImageGem(Ours) Cars Small & 0.2004 & \textbf{0.2741} & \textbf{0.3745} \\
            ImageGem(Ours) Cars Large & \textbf{0.2007} & 0.2738 & 0.3710 \\
            \midrule 
            Original SD1.5 & 0.2010 & 0.2646 & 0.3560 \\
            Pick-a-pic Dogs & 0.2058 & 0.2739 & 0.3617 \\
            ImageGem(Ours) Dogs & \textbf{0.2069} & \textbf{0.2789} & \textbf{0.3683} \\
            \midrule 
            Original SD1.5 & 0.1954 & 0.2640 & \textbf{0.3446} \\
            Pick-a-pic Scenery & 0.1936 & 0.2676 & 0.3289 \\
            ImageGem(Ours) Scenery Small & 0.1949 & 0.2730 & 0.3403 \\
            ImageGem(Ours) Scenery Large & \textbf{0.1961} & \textbf{0.2747} & 0.3427 \\ 
            \bottomrule
        \end{tabular}
    }
    \caption{Quantitative DiffusionDPO results comparing average scores: Pick Score~\cite{kirstain2023pick} and HPSv2~\cite{wu2023human} for human preference alignment, and CLIP Score~\cite{hessel2022clipscore} for image-prompt alignment.}
    \label{tab:diffusion_dpo}
\end{table}

We fine-tuned Stable Diffusion 1.5 (SD1.5) ~\cite{rombach2022high}, using three subsets sampled by specific key words shown in Tab.~\ref{tab:subsets}, from both our ImageGem and pick-a-pic~\cite{kirstain2023pick}. We use the original SD1.5 checkpoint, as well as the checkpoints fine-tuned with pick-a-pic subsets for baseline comparison. All checkpoints were trained with $4 \times$A100 GPUs, with batch size $1$ and gradient accumulation $128$ for $2000$ steps. The remaining hyperparameters were configured as described in DiffusionDPO~\cite{wallace2024diffusion}.
For evaluation, we sampled 200 Out-of-Distribution (OOD) prompts from DiffusionDB~\cite{wang2022diffusiondb} per topic and generated 600 images per checkpoint using three random seeds.

As shown in Tab.~\ref{tab:diffusion_dpo} and Fig.~\ref{fig:dpo_quali}, model checkpoints fine-tuned with subsets sampled from our datasets outperform those from Pick-a-Pic in all three topics. For the scenery topic, as we scale up the subset, improvements in all metrics are observed, but the CLIP score remains lower than the original SD1.5. We speculate that the DiffusionDPO training objective tends to prioritize models following human preference over prompt alignment.

\subsection{Retrieval and Generative Recommendation}
\label{sec:rec}
We sample another data subset for recommendation experiments, whose overall statistics are shown in Tab.~\ref{tab:rec-dataset}. For model recommendation, as user may interact with the same LoRA models multiple times, we select the last timestamp of the user interaction in our dataset.
We filter out users with less than 3 interactions for both image and model recommendation to eliminate cold-start scenarios.
For evaluation, we use leave-one-last ~\cite{meng2020exploring} with Recall@k and NDCG@k as retrieval and ranking metrics~\cite{jarvelin2002cumulated}, where Recall@k measures how often relevant items appear in the top-k list, and NDCG@k considers both presence and position, rewarding relevant items with higher ranking to assess ranking quality.

\subsubsection{Image and Model Retrieval}
For image retrieval, by explicitly modeling user interest as an embedding vector, the two-tower model yields the highest performance (Tab.~\ref{tab:image-rec}). For generative model retrieval, by introducing the structure of self-attention over user sequences, SASRec successfully captured temporal information in how each user's interest evolves, significantly outperforming traditional collaborative filtering methods such as ItemKNN and UserKNN (Tab.~\ref{tab:model-rec}). These results serve as baseline performances on our dataset, leaving the study of more sophisticated retrieval models for future work.

\begin{table}[t]
    \centering
    \scriptsize
    \setlength{\tabcolsep}{3pt}
    \renewcommand{\arraystretch}{0.9}
    \resizebox{\columnwidth}{!}{
        \begin{tabular}{lcccc}
            \toprule
            \textbf{Model} & \textbf{Rec@10000 $\uparrow$} & \textbf{Rec@5000 $\uparrow$} & \textbf{Rec@1000 $\uparrow$} & \textbf{Rec@100 $\uparrow$} \\
            \midrule
            ItemKNN & 0.4705 & 0.4190 & 0.3298 & 0.2342 \\
            Item2Vec & 0.5032 & 0.4425 & 0.3431 & 0.2399 \\
            Two-Tower & \textbf{0.5157} & \textbf{0.4479} & \textbf{0.3501} & \textbf{0.2402} \\
            \bottomrule
        \end{tabular}
    }
    \caption{Comparisons of retrieval performance on \textit{Images-1M}. ``Rec@k'' denotes Recall at rank $k$.}
    \label{tab:image-rec}
\end{table}

\begin{table}[t]
    \centering
    \scriptsize
    \setlength{\tabcolsep}{3pt}
    \renewcommand{\arraystretch}{0.9}
    \resizebox{\columnwidth}{!}{
        \begin{tabular}{lcccccc}
            \toprule
            \textbf{Model} & \textbf{Rec@100 $\uparrow$} & \textbf{Rec@50 $\uparrow$} & \textbf{Rec@10 $\uparrow$} & \textbf{NDCG@10 $\uparrow$} \\
            \midrule
            ItemKNN & 0.1282 & 0.1036 & 0.0773 & 0.057 \\
            UserKNN & 0.232 & 0.1818 & 0.1023 & 0.0705 \\
            SASRec & \textbf{0.2845} & \textbf{0.2451} & \textbf{0.1839} & \textbf{0.1239} \\
            \bottomrule
        \end{tabular}
    }
    \caption{Transposed comparison of ranking performance on \textit{Models-200K}. ``Rec@k'' denotes Recall at rank $k$.}
    \label{tab:model-rec}
\end{table}

\subsubsection{Generative Recommendation}
We randonly selected 20 users to test the feasibility of VLM-based ranking for generative recommendation. Each user’s test item, denoted as \( x_i \), appears in the top-10 retrieved list. To construct the VLM input for each user, we retain their latest \( H = 5 \) historical interactions and the top \( M = 10 \) retrieved items.

From Tab.~\ref{tab:combined_rec}, we observe that VLM shows promising potential in capturing user preference and items ranking in recommendation systems. In rankings of both image recommendation and model recommendation , VLM outperforms ItemKNN and SASRec in ranking quality. Compared to these traditional embedding-based methods, VLM provides human-readable and explainable rankings by generating textual justifications for its scores (Appendix~\ref{sec:appendix-example}). These results demonstrate that the captioning of user preferences by VLM effectively guides the ranking, improving both the ranking performance and the interpretability in recommendation systems. Furthermore, the success of VLM-based ranking highlights that structured textual representations can effectively capture user intent, which may further inspire advancements in generative recommendation tasks.

\begin{table}[t]
\centering
\small
\renewcommand{\arraystretch}{0.9}
\resizebox{\columnwidth}{!}{
\begin{tabular}{lcccc}
\toprule
\textbf{Method} & \textbf{Avg Rank $\downarrow$} & \textbf{Rank Std $\downarrow$} & \textbf{Rec@5 $\uparrow$} & \textbf{NDCG@5 $\uparrow$} \\
\midrule
\multicolumn{5}{c}{\textbf{Image Recommendation}} \\
\midrule
ItemKNN  & 2.9000  & 3.1439  & 0.7500  & \textbf{0.7065} \\
SASRec   & 3.4500  & 2.1145  & 0.8500  & 0.5494 \\
VLM      & \textbf{2.4000}  & \textbf{1.5355}  & \textbf{0.9500}  & 0.6745 \\
\midrule
\multicolumn{5}{c}{\textbf{Model Recommendation}} \\
\midrule
ItemKNN  & 7.9500  & 3.8179  & 0.2500  & 0.1509 \\
SASRec   & 5.3500  & \textbf{2.7004}  & 0.5000  & 0.2795 \\
VLM      & \textbf{3.9444}  & 2.7965  & \textbf{0.7222}  & \textbf{0.4981} \\
\bottomrule
\end{tabular}
}
\caption{Ranking performance of ItemKNN, SASRec, and VLM on image and model recommendation tasks.
}
\label{tab:combined_rec}
\end{table}

\subsection{Generative Model Personalization}
\subsubsection{Aggregated Preference Editing}
To assess the effectiveness of LoRA editing in the W2W space constructed with user-created LoRAs, we conducted a study to learn an editing direction from \textit{anime} to \textit{realistic} style, denoted as \textit{ani-real}, within the W2W space. This serves as a preliminary study to validate the W2W pipeline under a clear semantic shift before applying it to personalized preference alignment.

\textbf{Learning Tuning Direction.}  
We curated 23K SDXL-based LoRA models with diverse visual styles. Noting a predominance of human-focused styles in the model metadata, we chose to focus on learning an edit direction from \textit{anime} to \textit{realistic} within the human figure domain.
Initially, we used the ``tags'' field for binary labeling, but this approach proved noisy, as models tagged \textit{anime} often lacked anime characteristics in their sample outputs. 

To address this issue, we employed CLIP \cite{hessel2022clipscore} to compute the similarity between the models' example images and textual descriptions of the target styles. 
To build a reliable W2W space, we filtered models with a \textit{person} CLIP score $\geq$ 0.2. We then computed \textit{anime} and \textit{realistic} CLIP scores, excluding models with high values in both to avoid ambiguity. Among the rest, those with a \textit{realistic} CLIP score $\geq$ 0.26 were labeled 1, and those with an \textit{anime} score $\geq$ 0.24 were labeled -1, ensuring a clear editing direction from \textit{anime} to \textit{realistic} within the human figure domain.

\textbf{Limiting Input Size.}
\label{sec:limit_input_size}  
Applying Principal Component Analysis (PCA) to the weight space requires constraints on input matrix size. 
Given the variability in publicly generated LoRA models, we experiment with two alternative reduction strategies: (i) applying singular value decomposition (SVD), and (ii) selecting only LoRAs of a fixed rank and extracting specific layers.

Unlike the original W2W framework, which used self-trained rank-1 LoRAs \cite{dravid2024interpreting}, Civitai LoRAs vary in rank and structure. User-created LoRAs are often high-rank and vary widely in rank. To address this, we first experiment with input standardization by applying singular value decomposition (SVD) to each LoRA and retaining only the top-1 singular component (Appendix \ref{app:svd_prelim}). This significantly reduces weight size and ensures consistent dimensions across models, regardless of their original rank. As an alternative, we also experiment with selecting specific layers from fixed-rank LoRAs. We filtered for models with rank 16, yielding 857 LoRAs. To further reduce weight size, we selected only feed-forward (FF) and attention value (attn\_v) layers based on their significant impact on the base model (Appendix \ref{sec:different_layers_clip}). Since FF layers contain three times more parameters than attn\_v layers, we explored both to balance efficiency and effectiveness.

\begin{figure}[t]
    \centering
    \includegraphics[width=\columnwidth]{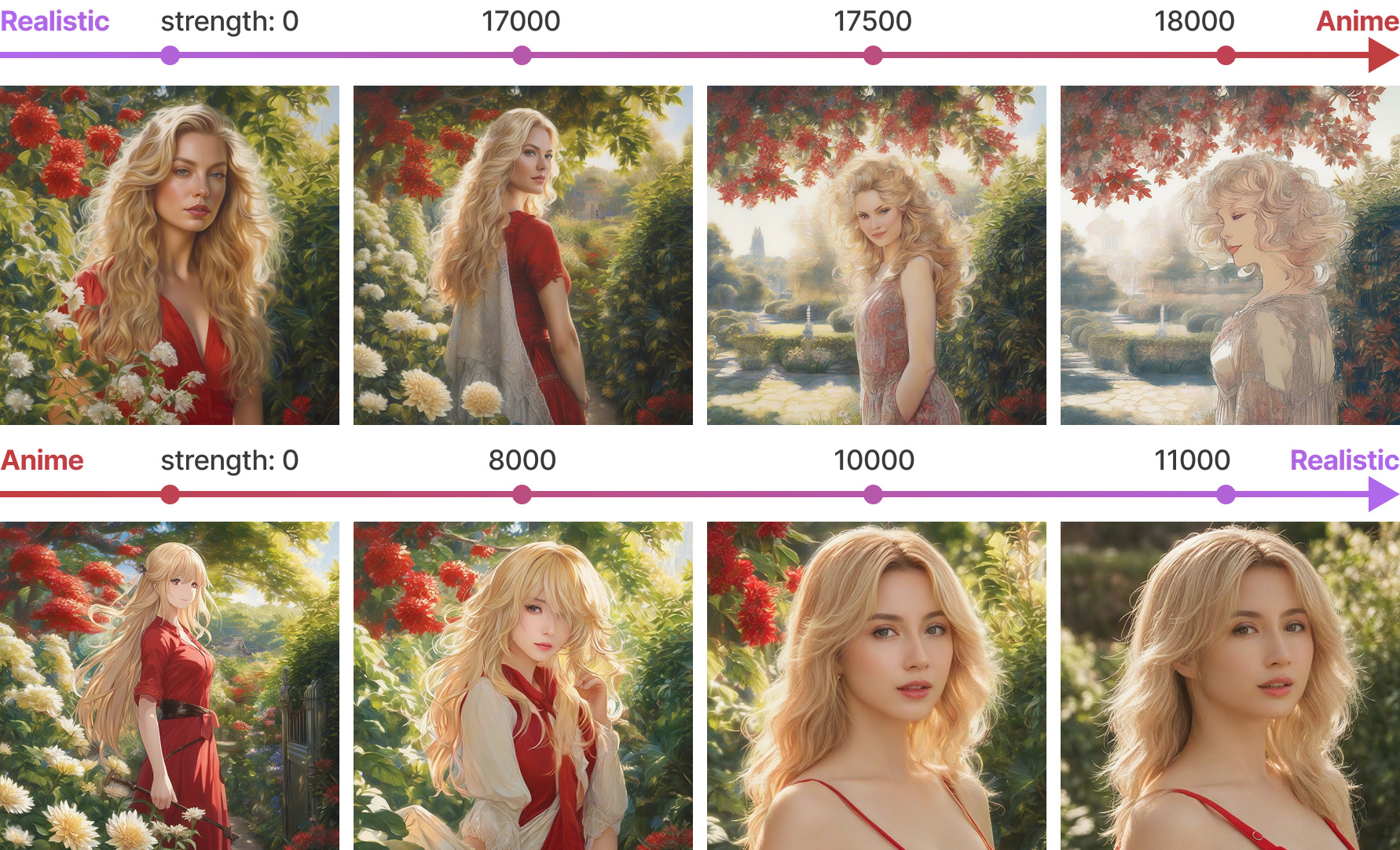}
    \captionsetup{width=\columnwidth}
    \caption{
        Perform tuning along the direction \textit{real-ani} (top) and direction \textit{ani-real} (bottom). The first column displays images generated by the unedited SDXL base model, while the subsequent columns show images generated by progressively edited models with increasing tuning strength. Each row shares the same generation seed for consistency. 
    }
    \label{fig:anti-real-weightspace}
\end{figure}

\textbf{Results.} 
Our results show that the SVD-based strategy yields the most robust transformations, enabling smooth and coherent edits in both the \textit{ani-real} and \textit{real-ani} directions (Figure~\ref{fig:anti-real-weightspace}). In contrast, the W2W space constructed from attn\_v layers performs well primarily in the \textit{ani-real} direction but fails to generalize to the reverse (Appendix \ref{sec:appendix_real-ani_results}). We also observe that using only FF layers leads to poor performance in both directions, suggesting that FF may not capture semantically aligned features necessary for effective editing. Since SVD operates on LoRAs of arbitrary rank and supports bidirectional, model-level editing with stronger consistency, we adopt it as primary approach.

\subsubsection{Individual Preference Learning}
Building upon the \textit{ani-real} transformation, we extend our approach to learn personalized editing directions within the W2W space in the human figure domain. 

\textbf{Preference Labeling.}
To capture individual preferences for user $P_i$, we compute CLIP embeddings for all their generated images, then apply HDBScan clustering to identify a representative preference cluster. 
To describe the user’s stylistic preference, we select the top-9 images closest to the cluster mean and use a Vision-Language Model (VLM) to generate textual descriptions of their common features. Following the approach in the \textit{ani-real} experiment, we compute CLIP similarity between LoRA models' example images and these descriptions. Models with higher similarity score are labeled 1, and those with lower similarity score are labeled -1, allowing us to learn a unique hyperplane to separate preferred and non-preferred models for each user.

\textbf{Multi-Direction LoRA Editing.}
We first demonstrate the effectiveness of learned preference directions by editing a single LoRA model $M_0$ that initially lies in the "not-preferred" region for two different users. Let $\vec{d_1}$ and $\vec{d_2}$ represent the learned editing directions for each user. We traverse the W2W space by updating  along each direction:\(M_1=M_0+\lambda_1\vec{d_1},M_2=M_0+\lambda_2\vec{d_2}\). 

\textbf{Multi-User Preference Alignment.}
To generalize preference learning, we select three users and construct three distinct preference directions $\left\{ \vec{d_i} \right\}_{i=1}^{3}$. For each direction, we choose two initial LoRA models $M_{i1}$ and $M_{i2}$, neither initially aligned with the respective user’s preference. We then update these models along their corresponding preference directions:\(M_{ij}' = M_{ij}+\lambda \vec{d_i}, j\in \left\{1,2\right\}\), where $M_{ij}'$ represents the preference-aligned model. For evaluation, we generate two images per model: before editing ($M_{ij}$) and after editing ($M_{ij}'$). We rank these images using CLIP and VLM: (1) We compute CLIP similarity between each image and the user preference cluster mean, ranking images by similarity; (2) We prompt VLM to compare each image to the user’s top-9 preferred images and rank by similarity.

\textbf{Results.}
As shown in Figure~\ref{fig:teaser}, the initial LoRA model $M_0$ lies in the "not-preferred" region for both users $P_1$ and $P_2$. By traversing the W2W space along their preference directions, we obtain two modified models, $M_1$ and $M_2$, that generate images better aligned with each user's stylistic preference.
Similarly, Figure~\ref{fig:multi_user} (Appendix \ref{sec:multi_user_appendix}) demonstrates preference alignment for three users, where initial misaligned models were adjusted along learned directions. Both CLIP and VLM rankings confirm improved alignment in images generated by the adjusted LoRA models. 
\section{Discussion}
Given the promising results from the three applications, we discuss several limitations and future work directions.

\textbf{Preference Data for DPO.}
We curated preference sets for DPO based on HPS~\cite{wu2023human} within each semantic cluster. Future experiments could explore ways to leverage the implicit feedback data from user interactions available in our dataset. Additionally, a possible extension is to conduct human preference alignment for larger, up-to-date diffusion models, such as Flux~\cite{flux2024}, leveraging the entire dataset.

\textbf{Generative Model Retrieval and Personalization.} 
We evaluated classical models across several retrieval and ranking tasks based on user interaction data, highlighting the space for improvement on generative model retrieval. 
Our dataset thus provides a testbed to further study this new task formulation in large-scale, where generative models are treated as ``items'' to be retrieved, and abundant user implicit feedback on prompts, tags and images are associated with these models.
We also presented a first look into the generative model personalization paradigm by directly editing a pre-trained LoRA according to user preference data. We demonstrated the promise of our approach from different image domains, but future work can explore how to generate models that align with various types of implicit user preference across multiple domains and data modalities.

\textbf{Constraint of PCA-based Weight Space.}
The reliance on PCA restricts model selection to low-rank (e.g., rank 8, rank 16), limiting the diversity of the models. Consequently, the available models within a given domain are constrained, which might lead to a lack of alignment with certain users' preferences. The limited number of models prohibits effective learning of W2W space for less popular domains (e.g., scenery). With a more diverse set of models, it would be possible to learn a wider range of meaningful directions in W2W space. Future work could explore alternative methods for learning LoRA weight spaces.

\section{Conclusion}
We propose ImageGem, a large-scale dataset consisting of in-the-wild user interactions with generative models and images. We show that our dataset empowers the study of various tasks related to preference alignment and personalization with generative models. We demonstrate for the first time a generative model personalization paradigm by customizing diffusion models in a latent weight space aligned with individual user preference. Our dataset opens a few new research directions on generative models for fine-grained preference learning and image generations.

\paragraph{Acknowledgement}
This work is supported in part by STCSM 23YF1430300 and NYU HPC resources.
\newpage
{
    \small
    \bibliographystyle{ieeenat_fullname}
    \bibliography{main}

\begin{thebibliography}{41}
\providecommand{\natexlab}[1]{#1}
\providecommand{\url}[1]{\texttt{#1}}
\expandafter\ifx\csname urlstyle\endcsname\relax
  \providecommand{\doi}[1]{doi: #1}\else
  \providecommand{\doi}{doi: \begingroup \urlstyle{rm}\Url}\fi

\bibitem[Agrawal et~al.(2024)Agrawal, Antoniak, Hanna, Bout, Chaplot, Chudnovsky, Costa, Monicault, Garg, Gervet, Ghosh, Héliou, Jacob, Jiang, Khandelwal, Lacroix, Lample, Casas, Lavril, Scao, Lo, Marshall, Martin, Mensch, Muddireddy, Nemychnikova, Pellat, Platen, Raghuraman, Rozière, Sablayrolles, Saulnier, Sauvestre, Shang, Soletskyi, Stewart, Stock, Studnia, Subramanian, Vaze, Wang, and Yang]{agrawal2024pixtral12b}
Pravesh Agrawal, Szymon Antoniak, Emma~Bou Hanna, Baptiste Bout, Devendra Chaplot, Jessica Chudnovsky, Diogo Costa, Baudouin~De Monicault, Saurabh Garg, Theophile Gervet, Soham Ghosh, Amélie Héliou, Paul Jacob, Albert~Q. Jiang, Kartik Khandelwal, Timothée Lacroix, Guillaume Lample, Diego~Las Casas, Thibaut Lavril, Teven~Le Scao, Andy Lo, William Marshall, Louis Martin, Arthur Mensch, Pavankumar Muddireddy, Valera Nemychnikova, Marie Pellat, Patrick~Von Platen, Nikhil Raghuraman, Baptiste Rozière, Alexandre Sablayrolles, Lucile Saulnier, Romain Sauvestre, Wendy Shang, Roman Soletskyi, Lawrence Stewart, Pierre Stock, Joachim Studnia, Sandeep Subramanian, Sagar Vaze, Thomas Wang, and Sophia Yang.
\newblock Pixtral 12b, 2024.

\bibitem[Barkan and Koenigstein(2016)]{barkan2016item2vec}
Oren Barkan and Noam Koenigstein.
\newblock Item2vec: neural item embedding for collaborative filtering.
\newblock In \emph{2016 IEEE 26th international workshop on machine learning for signal processing (MLSP)}, pages 1--6. IEEE, 2016.

\bibitem[Brooks et~al.(2022)Brooks, Holynski, and Efros]{Brooks2022InstructPix2PixLT}
Tim Brooks, Aleksander Holynski, and Alexei~A. Efros.
\newblock Instructpix2pix: Learning to follow image editing instructions.
\newblock \emph{2023 IEEE/CVF Conference on Computer Vision and Pattern Recognition (CVPR)}, pages 18392--18402, 2022.

\bibitem[Covington et~al.(2016)Covington, Adams, and Sargin]{covington2016deep}
Paul Covington, Jay Adams, and Emre Sargin.
\newblock Deep neural networks for youtube recommendations.
\newblock In \emph{Proceedings of the 10th ACM conference on recommender systems}, pages 191--198, 2016.

\bibitem[Damian0815(2024)]{compel}
Damian0815.
\newblock Compel, 2024.
\newblock GitHub repository, version 2.0.2.

\bibitem[Datta et~al.(2008)Datta, Joshi, Li, and Wang]{datta2008image}
Ritendra Datta, Dhiraj Joshi, Jia Li, and James~Z Wang.
\newblock Image retrieval: Ideas, influences, and trends of the new age.
\newblock \emph{ACM Computing Surveys (Csur)}, 40\penalty0 (2):\penalty0 1--60, 2008.

\bibitem[Deshpande and Karypis(2004)]{deshpande2004item}
Mukund Deshpande and George Karypis.
\newblock Item-based top-n recommendation algorithms.
\newblock \emph{ACM Transactions on Information Systems (TOIS)}, 22\penalty0 (1):\penalty0 143--177, 2004.

\bibitem[Dosovitskiy et~al.(2021)Dosovitskiy, Beyer, Kolesnikov, Weissenborn, Zhai, Unterthiner, Dehghani, Minderer, Heigold, Gelly, Uszkoreit, and Houlsby]{dosovitskiy2021an}
Alexey Dosovitskiy, Lucas Beyer, Alexander Kolesnikov, Dirk Weissenborn, Xiaohua Zhai, Thomas Unterthiner, Mostafa Dehghani, Matthias Minderer, Georg Heigold, Sylvain Gelly, Jakob Uszkoreit, and Neil Houlsby.
\newblock An image is worth 16x16 words: Transformers for image recognition at scale.
\newblock In \emph{International Conference on Learning Representations}, 2021.

\bibitem[Douze et~al.(2025)Douze, Guzhva, Deng, Johnson, Szilvasy, Mazaré, Lomeli, Hosseini, and Jégou]{douze2025faisslibrary}
Matthijs Douze, Alexandr Guzhva, Chengqi Deng, Jeff Johnson, Gergely Szilvasy, Pierre-Emmanuel Mazaré, Maria Lomeli, Lucas Hosseini, and Hervé Jégou.
\newblock The faiss library, 2025.

\bibitem[Dravid et~al.(2024)Dravid, Gandelsman, Wang, Abdal, Wetzstein, Efros, and Aberman]{dravid2024interpreting}
Amil Dravid, Yossi Gandelsman, Kuan-Chieh Wang, Rameen Abdal, Gordon Wetzstein, Alexei~A Efros, and Kfir Aberman.
\newblock Interpreting the weight space of customized diffusion models.
\newblock In \emph{The Thirty-eighth Annual Conference on Neural Information Processing Systems}, 2024.

\bibitem[Fan et~al.(2023)Fan, Watkins, Du, Liu, Ryu, Boutilier, Abbeel, Ghavamzadeh, Lee, and Lee]{fan2023dpok}
Ying Fan, Olivia Watkins, Yuqing Du, Hao Liu, Moonkyung Ryu, Craig Boutilier, Pieter Abbeel, Mohammad Ghavamzadeh, Kangwook Lee, and Kimin Lee.
\newblock Dpok: Reinforcement learning for fine-tuning text-to-image diffusion models.
\newblock \emph{Advances in Neural Information Processing Systems}, 36:\penalty0 79858--79885, 2023.

\bibitem[Gal et~al.(2023)Gal, Alaluf, Atzmon, Patashnik, Bermano, Chechik, and Cohen-or]{gal2023TextualInversion}
Rinon Gal, Yuval Alaluf, Yuval Atzmon, Or Patashnik, Amit~Haim Bermano, Gal Chechik, and Daniel Cohen-or.
\newblock An image is worth one word: Personalizing text-to-image generation using textual inversion.
\newblock In \emph{The Eleventh International Conference on Learning Representations}, 2023.

\bibitem[Guo et~al.(2024)Guo, Liu, and Wen]{guo2024gemrec}
Yuanhe Guo, Haoming Liu, and Hongyi Wen.
\newblock Gemrec: Towards generative model recommendation.
\newblock In \emph{Proceedings of the 17th ACM International Conference on Web Search and Data Mining}, 2024.

\bibitem[Hanu and {Unitary team}(2020)]{Detoxify}
Laura Hanu and {Unitary team}.
\newblock Detoxify.
\newblock Github. https://github.com/unitaryai/detoxify, 2020.

\bibitem[Hessel et~al.(2022)Hessel, Holtzman, Forbes, Bras, and Choi]{hessel2022clipscore}
Jack Hessel, Ari Holtzman, Maxwell Forbes, Ronan~Le Bras, and Yejin Choi.
\newblock Clipscore: A reference-free evaluation metric for image captioning, 2022.

\bibitem[Hu et~al.(2022)Hu, Shen, Wallis, Allen-Zhu, Li, Wang, Wang, Chen, et~al.]{hu2022lora}
Edward~J Hu, Yelong Shen, Phillip Wallis, Zeyuan Allen-Zhu, Yuanzhi Li, Shean Wang, Lu Wang, Weizhu Chen, et~al.
\newblock Lora: Low-rank adaptation of large language models.
\newblock \emph{ICLR}, 1\penalty0 (2):\penalty0 3, 2022.

\bibitem[J{\"a}rvelin and Kek{\"a}l{\"a}inen(2002)]{jarvelin2002cumulated}
Kalervo J{\"a}rvelin and Jaana Kek{\"a}l{\"a}inen.
\newblock Cumulated gain-based evaluation of ir techniques.
\newblock \emph{ACM Transactions on Information Systems (TOIS)}, 20\penalty0 (4):\penalty0 422--446, 2002.

\bibitem[Kang and McAuley(2018)]{kang2018selfattentivesequentialrecommendation}
Wang-Cheng Kang and Julian McAuley.
\newblock Self-attentive sequential recommendation, 2018.

\bibitem[Kirstain et~al.(2023)Kirstain, Polyak, Singer, Matiana, Penna, and Levy]{kirstain2023pick}
Yuval Kirstain, Adam Polyak, Uriel Singer, Shahbuland Matiana, Joe Penna, and Omer Levy.
\newblock Pick-a-pic: An open dataset of user preferences for text-to-image generation.
\newblock \emph{Advances in Neural Information Processing Systems}, 36:\penalty0 36652--36663, 2023.

\bibitem[Labs(2024)]{flux2024}
Black~Forest Labs.
\newblock Flux.
\newblock \url{https://github.com/black-forest-labs/flux}, 2024.

\bibitem[Liang et~al.(2024)Liang, He, Li, Li, Klimovskiy, Carolan, Sun, Pont-Tuset, Young, Yang, et~al.]{liang2024rich}
Youwei Liang, Junfeng He, Gang Li, Peizhao Li, Arseniy Klimovskiy, Nicholas Carolan, Jiao Sun, Jordi Pont-Tuset, Sarah Young, Feng Yang, et~al.
\newblock Rich human feedback for text-to-image generation.
\newblock In \emph{Proceedings of the IEEE/CVF Conference on Computer Vision and Pattern Recognition}, pages 19401--19411, 2024.

\bibitem[McInnes et~al.(2017)McInnes, Healy, and Astels]{McInnesLeland2017hHdb}
Leland McInnes, John Healy, and Steve Astels.
\newblock hdbscan: Hierarchical density based clustering.
\newblock \emph{Journal of open source software}, 2\penalty0 (11):\penalty0 205--, 2017.

\bibitem[McInnes et~al.(2018)McInnes, Healy, Saul, and Großberger]{McInnes2018UMAP}
Leland McInnes, John Healy, Nathaniel Saul, and Lukas Großberger.
\newblock Umap: Uniform manifold approximation and projection.
\newblock \emph{Journal of Open Source Software}, 3\penalty0 (29):\penalty0 861, 2018.

\bibitem[Meng et~al.(2020)Meng, McCreadie, Macdonald, and Ounis]{meng2020exploring}
Zaiqiao Meng, Richard McCreadie, Craig Macdonald, and Iadh Ounis.
\newblock Exploring data splitting strategies for the evaluation of recommendation models.
\newblock In \emph{Proceedings of the 14th acm conference on recommender systems}, pages 681--686, 2020.

\bibitem[Podell et~al.(2024)Podell, English, Lacey, Blattmann, Dockhorn, M{\"u}ller, Penna, and Rombach]{podell2024sdxl}
Dustin Podell, Zion English, Kyle Lacey, Andreas Blattmann, Tim Dockhorn, Jonas M{\"u}ller, Joe Penna, and Robin Rombach.
\newblock {SDXL}: Improving latent diffusion models for high-resolution image synthesis.
\newblock In \emph{The Twelfth International Conference on Learning Representations}, 2024.

\bibitem[Qian et~al.(2024)Qian, Wang, Patashnik, Heravi, Ostashev, Tulyakov, Cohen-Or, and Aberman]{qian2024omni}
Guocheng Qian, Kuan-Chieh Wang, Or Patashnik, Negin Heravi, Daniil Ostashev, Sergey Tulyakov, Daniel Cohen-Or, and Kfir Aberman.
\newblock Omni-id: Holistic identity representation designed for generative tasks.
\newblock \emph{arXiv preprint arXiv:2412.09694}, 2024.

\bibitem[Radford et~al.(2021)Radford, Kim, Hallacy, Ramesh, Goh, Agarwal, Sastry, Askell, Mishkin, Clark, Krueger, and Sutskever]{Radford2021CLIP}
Alec Radford, Jong~Wook Kim, Chris Hallacy, Aditya Ramesh, Gabriel Goh, Sandhini Agarwal, Girish Sastry, Amanda Askell, Pamela Mishkin, Jack Clark, Gretchen Krueger, and Ilya Sutskever.
\newblock Learning transferable visual models from natural language supervision.
\newblock In \emph{International Conference on Machine Learning}, 2021.

\bibitem[Rombach et~al.(2022)Rombach, Blattmann, Lorenz, Esser, and Ommer]{rombach2022high}
Robin Rombach, Andreas Blattmann, Dominik Lorenz, Patrick Esser, and Bj{\"o}rn Ommer.
\newblock High-resolution image synthesis with latent diffusion models.
\newblock In \emph{Proceedings of the IEEE/CVF conference on computer vision and pattern recognition}, pages 10684--10695, 2022.

\bibitem[Ruiz et~al.(2023)Ruiz, Li, Jampani, Pritch, Rubinstein, and Aberman]{ruiz2023dreambooth}
Nataniel Ruiz, Yuanzhen Li, Varun Jampani, Yael Pritch, Michael Rubinstein, and Kfir Aberman.
\newblock Dreambooth: Fine tuning text-to-image diffusion models for subject-driven generation.
\newblock In \emph{Proceedings of the IEEE/CVF conference on computer vision and pattern recognition}, pages 22500--22510, 2023.

\bibitem[Ruiz et~al.(2024)Ruiz, Li, Jampani, Wei, Hou, Pritch, Wadhwa, Rubinstein, and Aberman]{ruiz2024hyperdreambooth}
Nataniel Ruiz, Yuanzhen Li, Varun Jampani, Wei Wei, Tingbo Hou, Yael Pritch, Neal Wadhwa, Michael Rubinstein, and Kfir Aberman.
\newblock Hyperdreambooth: Hypernetworks for fast personalization of text-to-image models.
\newblock In \emph{Proceedings of the IEEE/CVF conference on computer vision and pattern recognition}, pages 6527--6536, 2024.

\bibitem[Salehi et~al.(2024)Salehi, Shafiei, Yeo, Bachmann, and Zamir]{salehi2024viper}
Sogand Salehi, Mahdi Shafiei, Teresa Yeo, Roman Bachmann, and Amir Zamir.
\newblock Viper: Visual personalization of generative models via individual preference learning.
\newblock In \emph{European Conference on Computer Vision}, pages 391--406. Springer, 2024.

\bibitem[Wallace et~al.(2024)Wallace, Dang, Rafailov, Zhou, Lou, Purushwalkam, Ermon, Xiong, Joty, and Naik]{wallace2024diffusion}
Bram Wallace, Meihua Dang, Rafael Rafailov, Linqi Zhou, Aaron Lou, Senthil Purushwalkam, Stefano Ermon, Caiming Xiong, Shafiq Joty, and Nikhil Naik.
\newblock Diffusion model alignment using direct preference optimization.
\newblock In \emph{Proceedings of the IEEE/CVF Conference on Computer Vision and Pattern Recognition}, pages 8228--8238, 2024.

\bibitem[Wang et~al.(2024)Wang, Bai, Wang, Qin, and Chen]{wang2024instantid}
Qixun Wang, Xu Bai, Haofan Wang, Zekui Qin, and Anthony Chen.
\newblock Instantid: Zero-shot identity-preserving generation in seconds.
\newblock \emph{arXiv preprint arXiv:2401.07519}, 2024.

\bibitem[Wang et~al.(2022)Wang, Montoya, Munechika, Yang, Hoover, and Chau]{wang2022diffusiondb}
Zijie~J Wang, Evan Montoya, David Munechika, Haoyang Yang, Benjamin Hoover, and Duen~Horng Chau.
\newblock Diffusiondb: A large-scale prompt gallery dataset for text-to-image generative models.
\newblock \emph{arXiv preprint arXiv:2210.14896}, 2022.

\bibitem[Wang et~al.(2023)Wang, Hohman, and Chau]{wangWizMapScalableInteractive2023}
Zijie~J. Wang, Fred Hohman, and Duen~Horng Chau.
\newblock {{WizMap}}: {{Scalable Interactive Visualization}} for {{Exploring Large Machine Learning Embeddings}}.
\newblock \emph{arXiv 2306.09328}, 2023.

\bibitem[Wu et~al.(2025)Wu, Xu, Po, Zhang, Yang, Wang, Liu, Lin, and Wetzstein]{wu2025fiva}
Tong Wu, Yinghao Xu, Ryan Po, Mengchen Zhang, Guandao Yang, Jiaqi Wang, Ziwei Liu, Dahua Lin, and Gordon Wetzstein.
\newblock Fiva: Fine-grained visual attribute dataset for text-to-image diffusion models.
\newblock \emph{Advances in Neural Information Processing Systems}, 37:\penalty0 31990--32011, 2025.

\bibitem[Wu et~al.(2023)Wu, Sun, Zhu, Zhao, and Li]{wu2023human}
Xiaoshi Wu, Keqiang Sun, Feng Zhu, Rui Zhao, and Hongsheng Li.
\newblock Human preference score: Better aligning text-to-image models with human preference.
\newblock In \emph{Proceedings of the IEEE/CVF International Conference on Computer Vision}, pages 2096--2105, 2023.

\bibitem[Xu et~al.(2023)Xu, Liu, Wu, Tong, Li, Ding, Tang, and Dong]{xu2023imagereward}
Jiazheng Xu, Xiao Liu, Yuchen Wu, Yuxuan Tong, Qinkai Li, Ming Ding, Jie Tang, and Yuxiao Dong.
\newblock Imagereward: Learning and evaluating human preferences for text-to-image generation.
\newblock \emph{Advances in Neural Information Processing Systems}, 36:\penalty0 15903--15935, 2023.

\bibitem[Yi et~al.(2019)Yi, Yang, Hong, Cheng, Heldt, Kumthekar, Zhao, Wei, and Chi]{yi2019sampling}
Xinyang Yi, Ji Yang, Lichan Hong, Derek~Zhiyuan Cheng, Lukasz Heldt, Aditee Kumthekar, Zhe Zhao, Li Wei, and Ed Chi.
\newblock Sampling-bias-corrected neural modeling for large corpus item recommendations.
\newblock In \emph{Proceedings of the 13th ACM conference on recommender systems}, pages 269--277, 2019.

\bibitem[Zhang et~al.(2023)Zhang, Rao, and Agrawala]{Zhang2023ControlNet}
Lvmin Zhang, Anyi Rao, and Maneesh Agrawala.
\newblock Adding conditional control to text-to-image diffusion models.
\newblock \emph{2023 IEEE/CVF International Conference on Computer Vision (ICCV)}, pages 3813--3824, 2023.

\bibitem[Zhang et~al.(2024)Zhang, Huang, Chen, Zhang, Wu, Feng, Wang, Shen, Liu, and Luo]{zhang2024flashface}
Shilong Zhang, Lianghua Huang, Xi Chen, Yifei Zhang, Zhi-Fan Wu, Yutong Feng, Wei Wang, Yujun Shen, Yu Liu, and Ping Luo.
\newblock Flashface: Human image personalization with high-fidelity identity preservation.
\newblock \emph{arXiv preprint arXiv:2403.17008}, 2024.

\end{thebibliography}
}
\clearpage
\appendix

\section{Dataset}
\subsection{Image Prompts Safety Check}

\begin{figure}[h]
    \centering
    \includegraphics[width=0.48\textwidth]{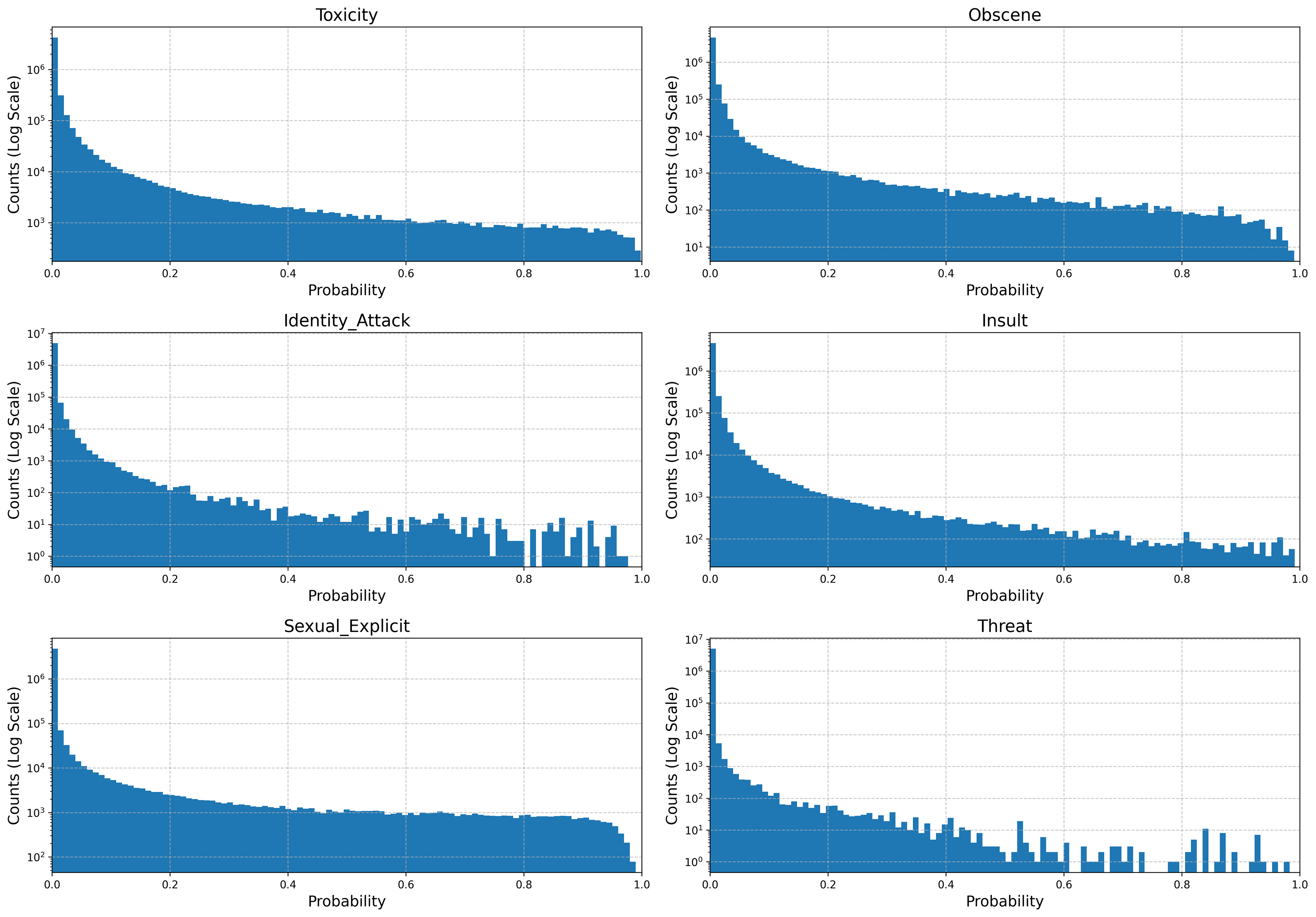}
    \caption{The distribution of prompts based on their predicted probabilites for NSFW content using Detoxify~\cite{Detoxify}. The y-axis represents the count of propmts in logarithmic scale.}
    \label{fig:safety_prompt}
\end{figure}

Fig.~\ref{fig:safety_prompt} shows the predicted the probability of NSFW content with Detoxify~\cite{Detoxify}
for six aspects: toxicity, obscenity, identity attack, insult, threat, and sexual explicitness. 

\subsection{Dataset Details}

\begin{figure}[h]
    \centering
    \includegraphics[width=0.48\textwidth]{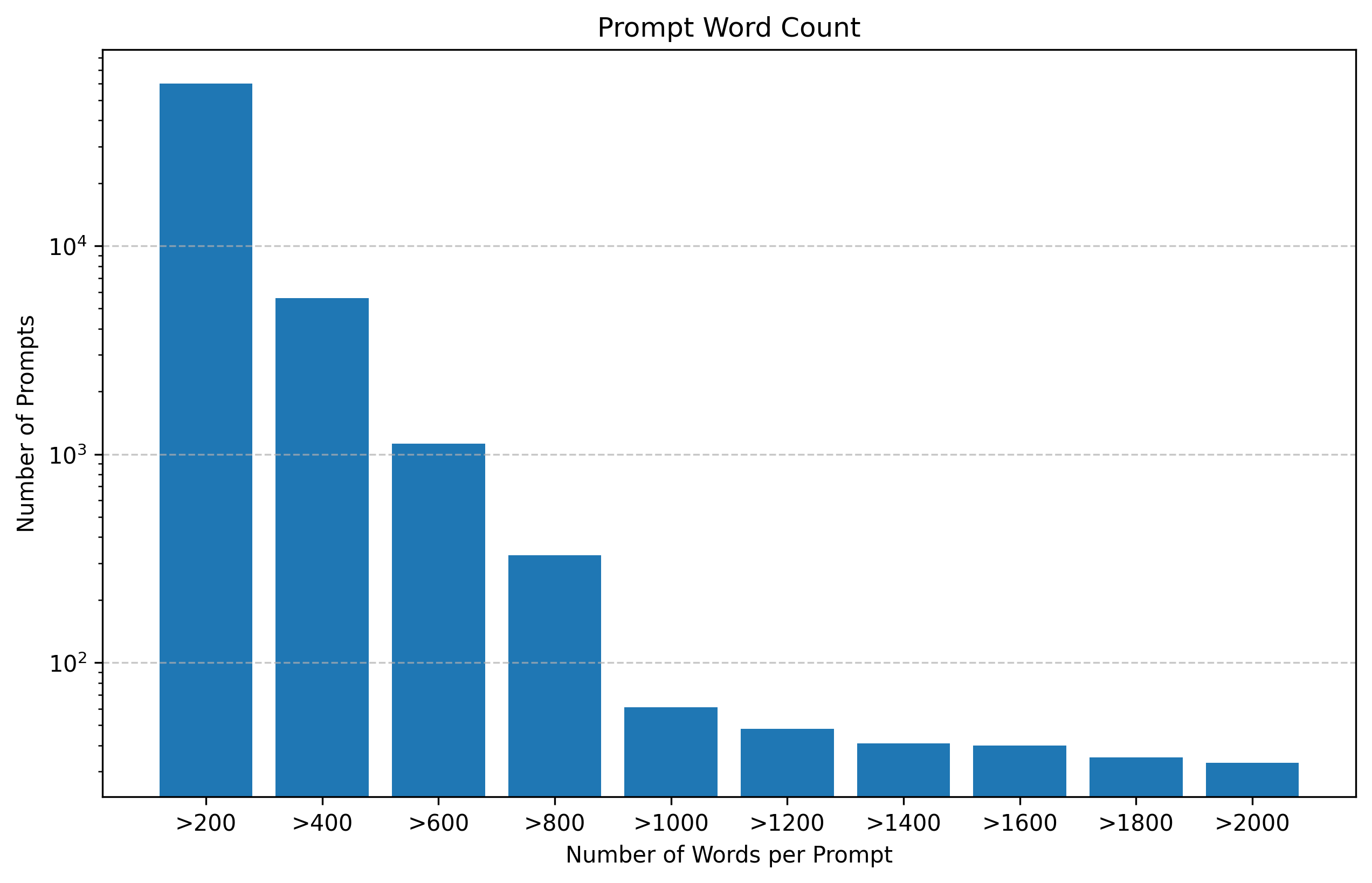}
    \captionsetup{width=\columnwidth}
    \caption{
        Cumulative Distribution of Prompt Word Counts in Log Scale for Prompts Exceeding 200 Words.
    }
    \label{fig:promp_word_count}
\end{figure}

\textbf{Prompt Word Count.} We observed some exceptionally long prompts in our dataset. Fig.~\ref{fig:promp_word_count} shows the distribution of word counts for prompts with more than $200$ words.

\begin{figure}[h!]
    \centering
    \includegraphics[width=0.48\textwidth]{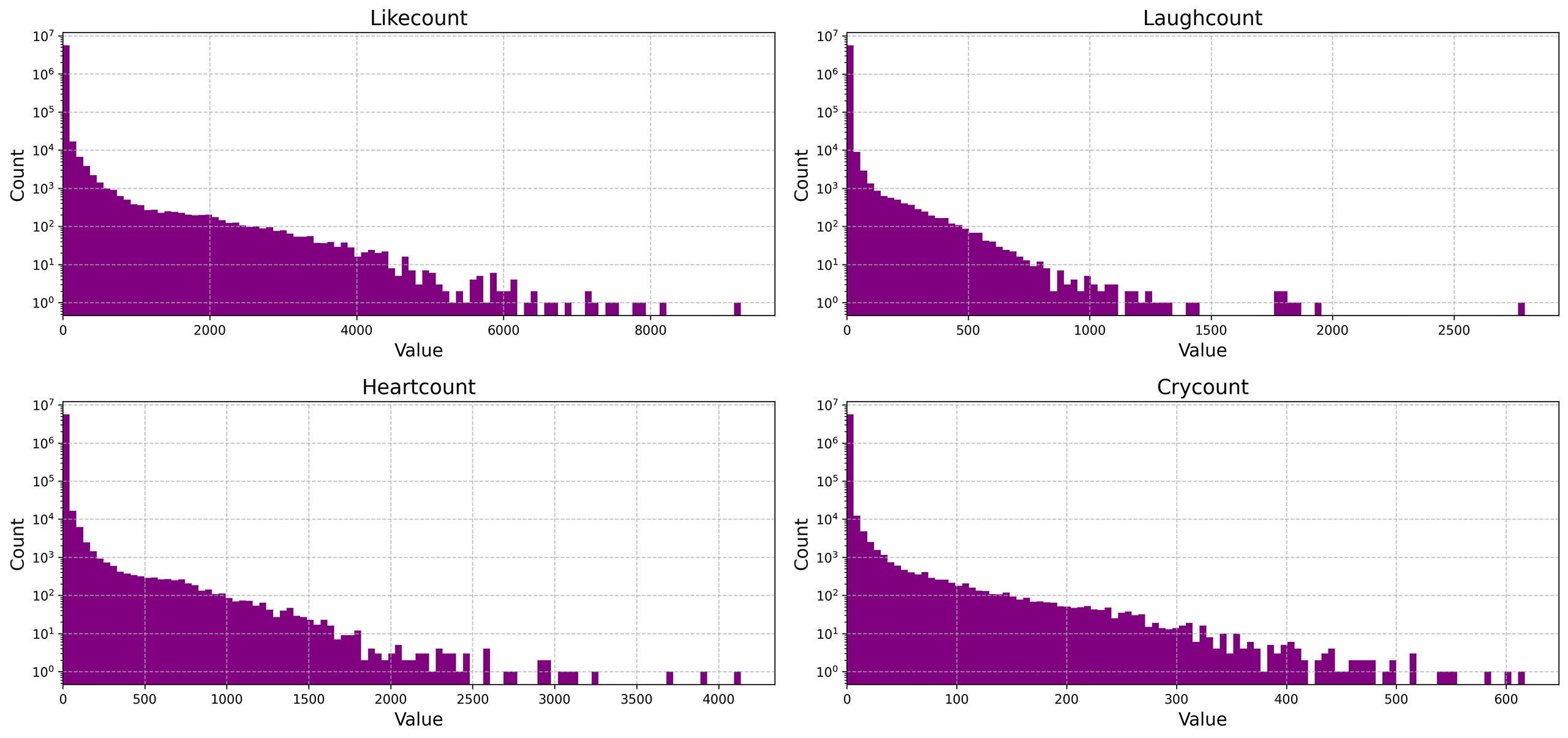}
    \caption{Log-scale distribution of image interactions for each emoji, with interaction values on the x-axis and the number of images on the y-axis.}
    \label{fig:image_stats}
\end{figure}

\textbf{User-Image Feedback.} 
Civitai enables users to respond to images with emojis anonymously, including ``Heart'', ``Like'' (Thumbs Up), ``Laugh'', ``cry''. Fig.~\ref{fig:image_stats} shows the distribution of these user-image interactions, which could serve as an indicator of popularity biases.

\begin{figure}[h]
    \centering
    \includegraphics[width=0.48\textwidth]{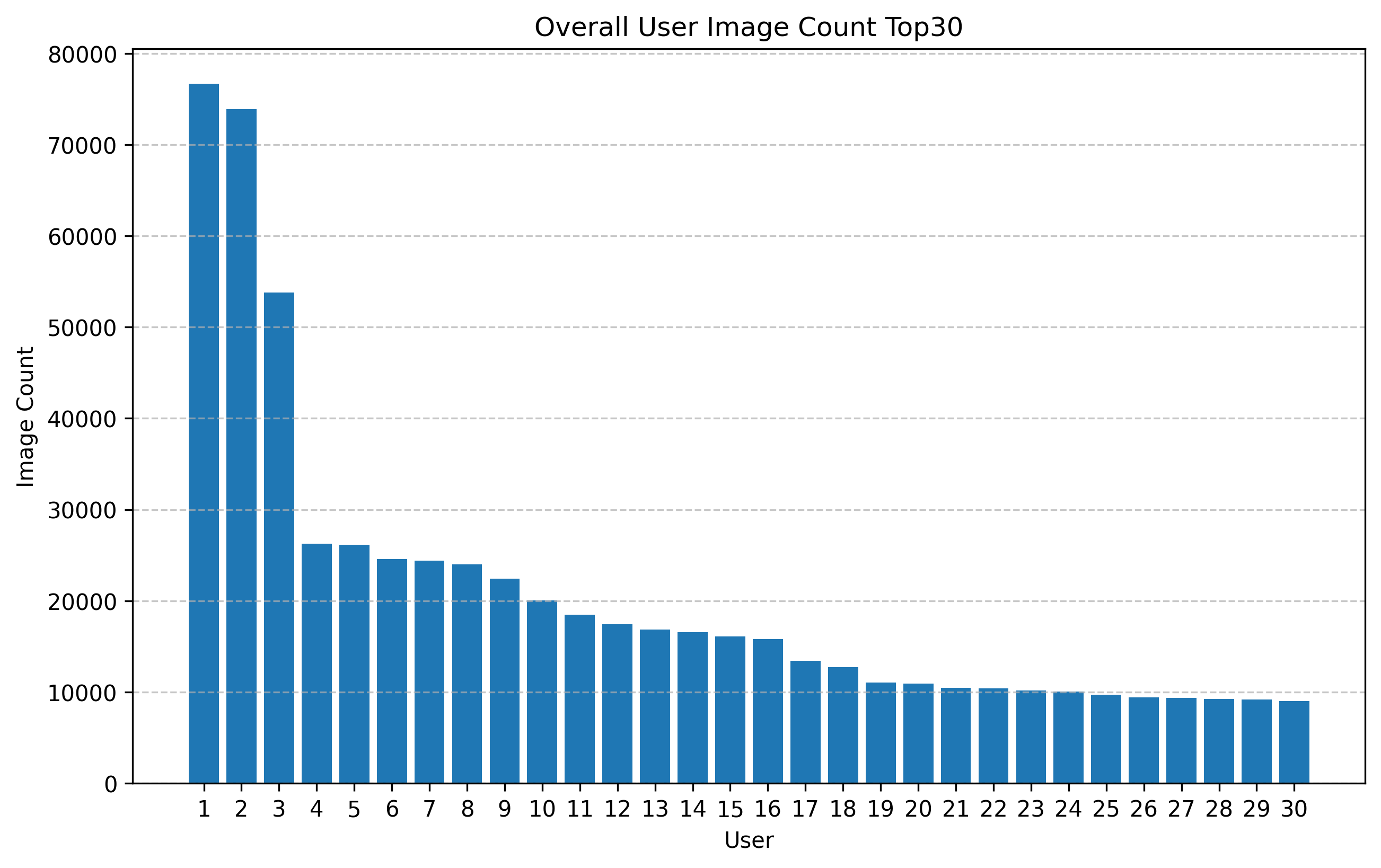}
    \captionsetup{width=\columnwidth}
    \caption{
        Top 30 uses based on their image count. User names are hidden for privacy.
    }
    \label{fig:user_image}
\end{figure}

\begin{figure}[h!]
    \centering
    \includegraphics[width=0.48\textwidth]{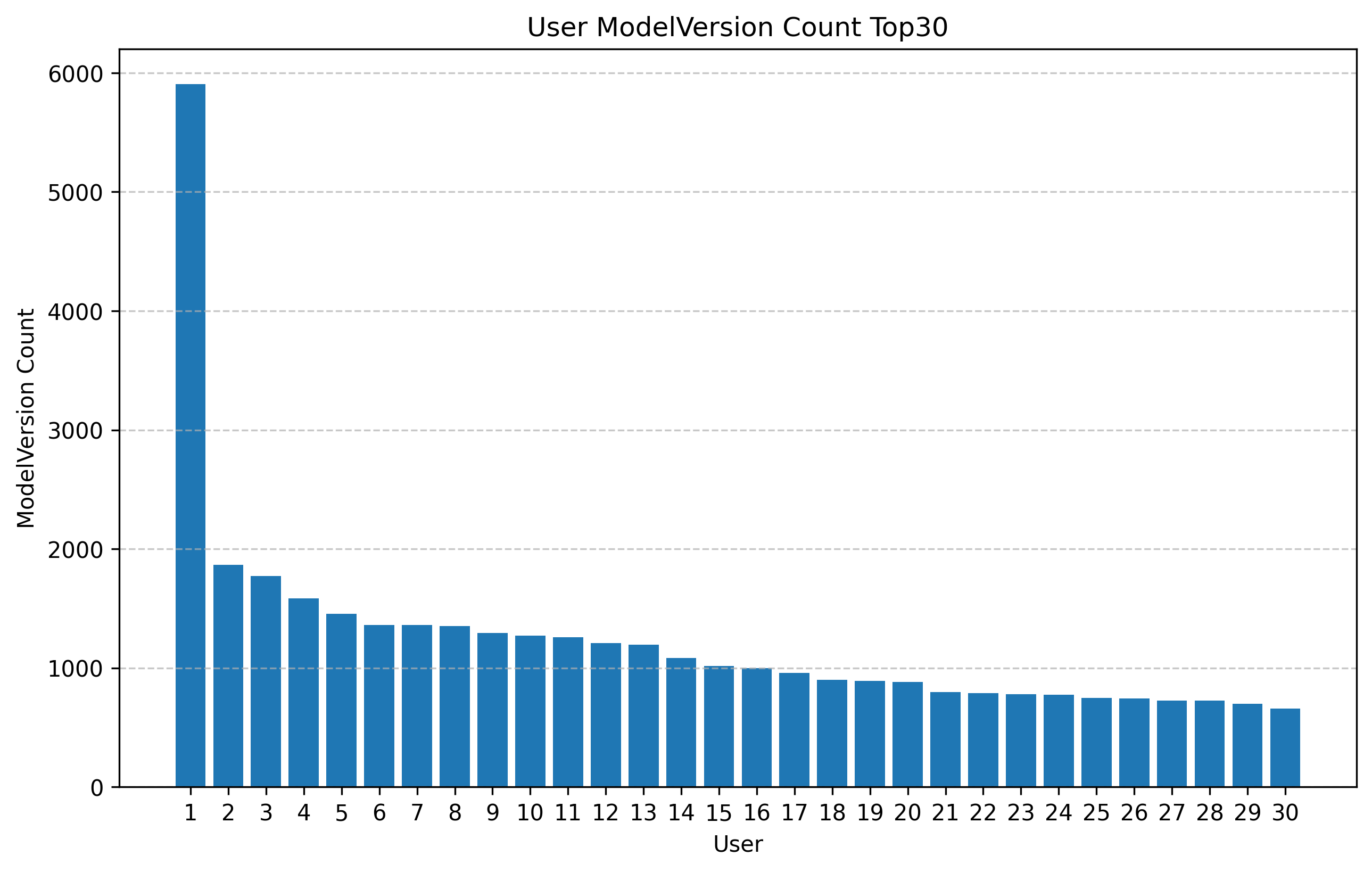}
    \captionsetup{width=\columnwidth}
    \caption{
        Top 30 users based on their model checkpoint count. User names are hidden for privacy.
    }
    \label{fig:user_modelversion}
\end{figure}

\textbf{User Interactions.}
We observe that the distribution of both user-image interactions and user-model interaction follows a long-tail manner. Fig.~\ref{fig:user_image} plots the top30 users for image count and Fig.~\ref{fig:user_modelversion} shows the top30 users for model count.

\section{VLM Captioning and Ranking}
\label{sec:appendix-vlm-ranking}
\subsection{VLM Prompting Strategies and Ranking Demonstration}
\label{sec:appendix-prompts}

This appendix presents the structured prompts used in our VLM recommendation system, categorized into image recommendation and model recommendation, each with captioning and ranking tasks.

\subsubsection{Image Captioning}

\begin{lstlisting}
Analyze these images and generate a structured description focusing on:

1. Primary Subject Type (e.g., human, fantasy creature, landscape).

2. Defining Visual Features (facial structure, clothing details, body posture).

3. Artistic Style (anime, realistic, digital painting).

4. Background Elements (futuristic city, ancient palace, foggy forest).
\end{lstlisting} 
\subsubsection{Image Ranking}
\begin{lstlisting}
Rank images based on similarity to the visual preference profile.

1. Overall Similarity (60 pts)
   - Primary Subject Match (20 pts): Does it belong to the same category? (Human, anthropomorphic, animal, scenery, object)
   - Artistic Style (15 pts): Matches reference? (Anime, realistic, digital painting, etc.)
   - Color Palette & Mood (15 pts): Similar tones, lighting, contrast?
   - Background & Setting (10 pts): Same environment (indoor, nature, fantasy, city, etc.)?

2. Detail Similarity (40 pts)
   - Key Features (20 pts):
       - Humans: Hair, clothing, accessories.
       - Animals: Fur color, body shape, eye design.
       - Scenery/Objects: Texture, materials, lighting effects.
   - Pose & Expression (10 pts): Consistency with visual preference profile.
   - Fine Details (10 pts): Composition, small artistic elements.

Return a JSON object:

{
    "image\_id": ID, 
    "similarity\_score": score,
    "explanation": "Brief reason" 
}
\end{lstlisting}

\subsubsection{Model Captioning}
\begin{lstlisting}
Summarize the common features, themes, and styles across these descriptions in detail.
\end{lstlisting}
\subsubsection{Model Ranking}
\begin{lstlisting}
Extract a detailed description of the user's visual style preferences.

Compare prompts based on:

1. Primary Subject (e.g., architecture, people, nature, abstract).

2. Artistic Style & Features (e.g., brushwork, realism, shading).

3. Color, Composition, Lighting (e.g., soft pastels, dark cyberpunk, 
contrast).

Scoring:

90-100: Perfect match with all key preferences
70-89: Strong match with most preferences
50-69: Moderate match with some preferences
30-49: Weak match with few preferences
10-29: Very weak match with preferences
0-9: No match with preferences

Return a JSON object:
{
    "version\_id": Version ID,
    "similarity\_score": score,  
    "explanation": "Brief reason" 
}
\end{lstlisting}

\subsubsection{Randomized Scoring Strategy}
To address the instability of VLM ranking results, we randomly sample a subset \( C_i^{(k)} \subseteq C_i \) of \( k \) items, repeat the VLM scoring process \( T \) times with different sampled subsets, and compute the final score \( s(x) \) for each item \( x \in C_i \) as the expectation over multiple trials.
This strategy ensures more consistent evaluations rather than relying on a single inference pass.

\subsection{Example of VLM Ranking}
\label{sec:appendix-example}

\begin{figure*}[htbp]
    \centering
    \includegraphics[width=1\textwidth]{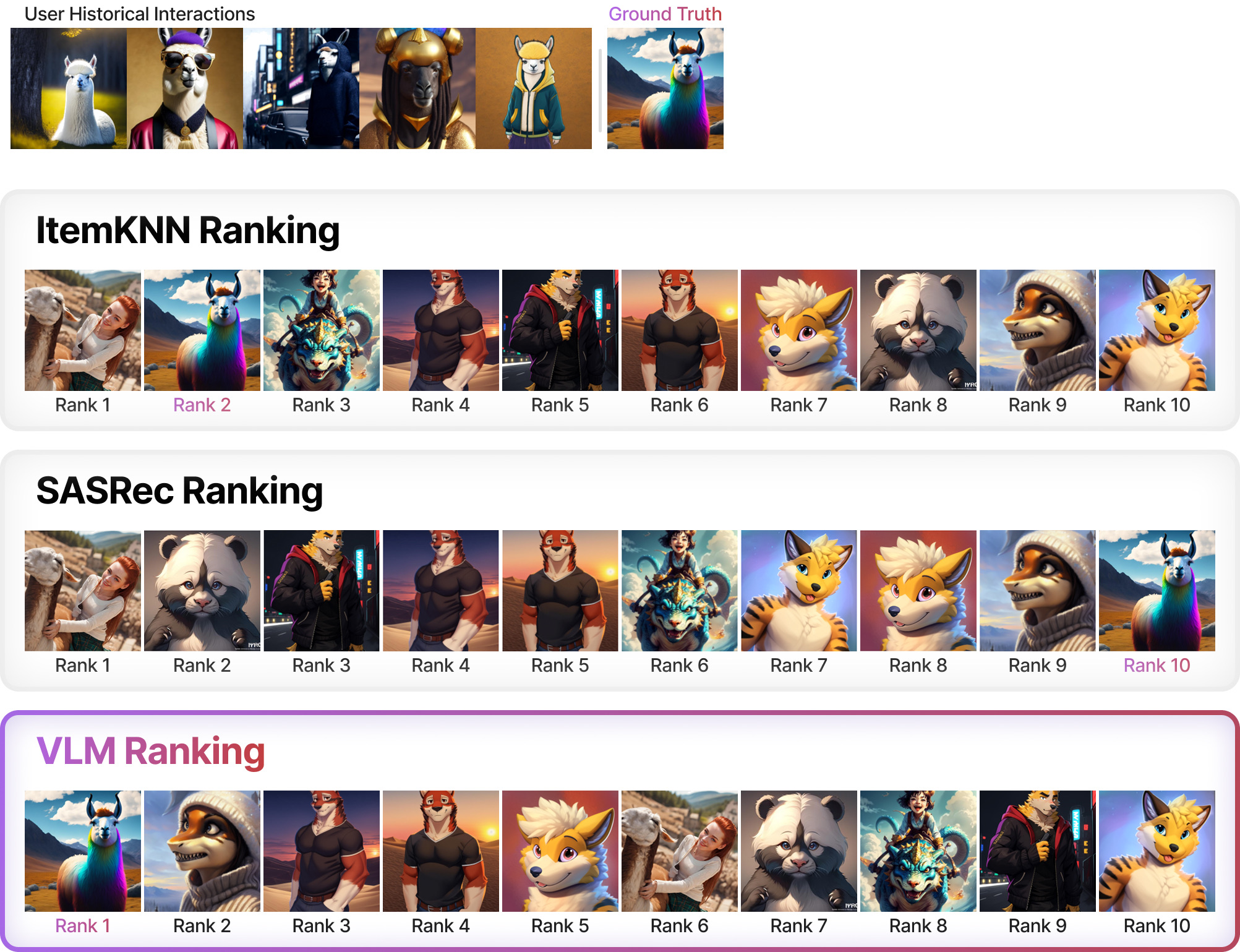} 
    \caption{The top row represents the user's historical interactions (training set). The following rows show rankings from three recommendation models: ItemKNN, SASRec, and VLM. Images are ordered by ranking from left to right. The VLM model demonstrates superior performance, as its rankings align most closely with the user's ground truth interaction.}
    \label{fig:ranking_pic}
\end{figure*}

The Table~\ref{tab:vlm_ranking} and Figure~\ref{fig:ranking_pic} presents VLM ranking results from the same user. Table~\ref{tab:vlm_ranking} presents the ranked images along with their similarity scores and explanations. These rankings correspond directly to the visual results in Figure~\ref{fig:ranking_pic}, demonstrating VLM's interpretability—each ranked image is accompanied by a justification. Additionally, the ground truth (GT) image is ranked relatively high, showcasing VLM's promising performance. This example further illustrates how VLM-generated user preferences effectively guide ranking, contributing to more personalized and explainable recommendations.

\begin{table*}[htbp]
\centering
\small
\setlength{\tabcolsep}{6pt}
\renewcommand{\arraystretch}{1.2}
\resizebox{\textwidth}{!}{  
\begin{tabular}{lcl}
\toprule
\textbf{Image ID} & \textbf{Similarity Score} & \textbf{Explanation} \\
\midrule
242811 & 82.5 & High similarity with primary subject match, artistic style, color palette, and key facial features.\\
173182 & 79.5 & Good match with similar facial features and similar anime style. \\
660727 & 78.3 & High similarity with key features, but difference in clothing and background. \\
244921 & 76.3 & Decent match with feminine features but less intricate in background details. \\
244821 & 76.0 & High overall similarity, similar style and key features but slight difference in color palette. \\
173226 & 72.7 & Moderate match with some preferences but weaker in details and artistic style compared to the highest matches. \\
173227 & 70.0 & Moderate similarity with key features but significant difference in style and color palette. \\
456861 & 69.3 & Weak match with key preferences; differences in artistic style, color palette, and less pronounced facial features. \\
523827 & 68.3 & Moderate match overall, slightly weaker because of hybrid eye color and differences in artistic style and setting. \\
456856 & 62.7 & Weak match due to differences in artistic style, background, and slight disparity in key facial features. \\
\bottomrule
\end{tabular}
}
\caption{VLM assigns higher scores to images that closely match key visual features. Lower-ranked images often exhibit differences in background details, artistic style, or facial attributes, highlighting VLM’s ability to provide an interpretable ranking explanation.}
\label{tab:vlm_ranking}
\end{table*}

\FloatBarrier

\FloatBarrier
\cleardoublepage

\section{Generative Model Personalization}

\subsection{SVD Preliminary Study}
\label{app:svd_prelim}

To evaluate the effectiveness of SVD-based rank reduction, we decompose each LoRA into singular vectors and retain only the top-1 component. Using the same seed and prompt, we generate images from three models: the base SDXL model, the user-created full-rank LoRA, and the corresponding rank-1 reduced LoRA. We compute CLIP similarity between the base model's image and each LoRA-generated image to assess fidelity. As shown in Tab.~\ref{tab:svd_clip}, rank-1 LoRA shows only a slight increase in average CLIP similarity compared to the full-rank version, suggesting that the top-1 singular direction captures most of the useful information. This experiment is conducted across 10178 SDXL LoRAs with an average rank of 23.95.

\begin{table}[htbp]
    \centering
    \scriptsize
    \setlength{\tabcolsep}{3pt}
    \resizebox{0.45\textwidth}{!}{
    \begin{tabular}{lcc}
        \toprule
        \textbf{Model Type} & \textbf{Avg. CLIP Score} & \textbf{Std Dev} \\
        \midrule
        Rank-1 LoRA & 0.8114 & 0.1151 \\
        Full-Rank LoRA & 0.7563 & 0.1215 \\
        \bottomrule
    \end{tabular}
    }
    \caption{CLIP similarity between images generated by the unedited SDXL base model and those generated using the original high-rank LoRA and its SVD-reduced rank-1 version.}
    \label{tab:svd_clip}
\end{table}

\subsection{Significance of Different Layers}
\label{sec:different_layers_clip}
To assess the significance of different LoRA layers, we conducted experiments by injecting weight residuals from individual layers into a base model. Using identical seeds, we generated images and computed CLIP scores to measure the difference between these images and those from the base model. The results in Tab. \ref{tab:clip_scores} showed that feed-forward (FF) and attention value (attn\_v) layers had the most significant impact on image generation

\begin{table}[t]
    \centering
    \scriptsize
    \setlength{\tabcolsep}{3pt}
    \resizebox{0.35\textwidth}{!}{
    \begin{tabular}{lc}
        \toprule
        \textbf{Layer Type} & \textbf{Average CLIP Score} \\
        \midrule
        attn\_v        & 0.8851  \\
        attn           & 0.8433  \\
        ff             & 0.8319  \\
        ff+attn\_v     & 0.7774  \\
        \bottomrule
    \end{tabular}
    }
    \caption{Comparison of CLIP scores across different layer types. Scores are averaged over 24 models.}
    \label{tab:clip_scores}
\end{table}

\subsection{\textit{ani-real} and \textit{real-ani} Editing Results}
\label{sec:appendix_real-ani_results}
\begin{figure*}[ht]
    \centering
    \includegraphics[width=0.85\linewidth]{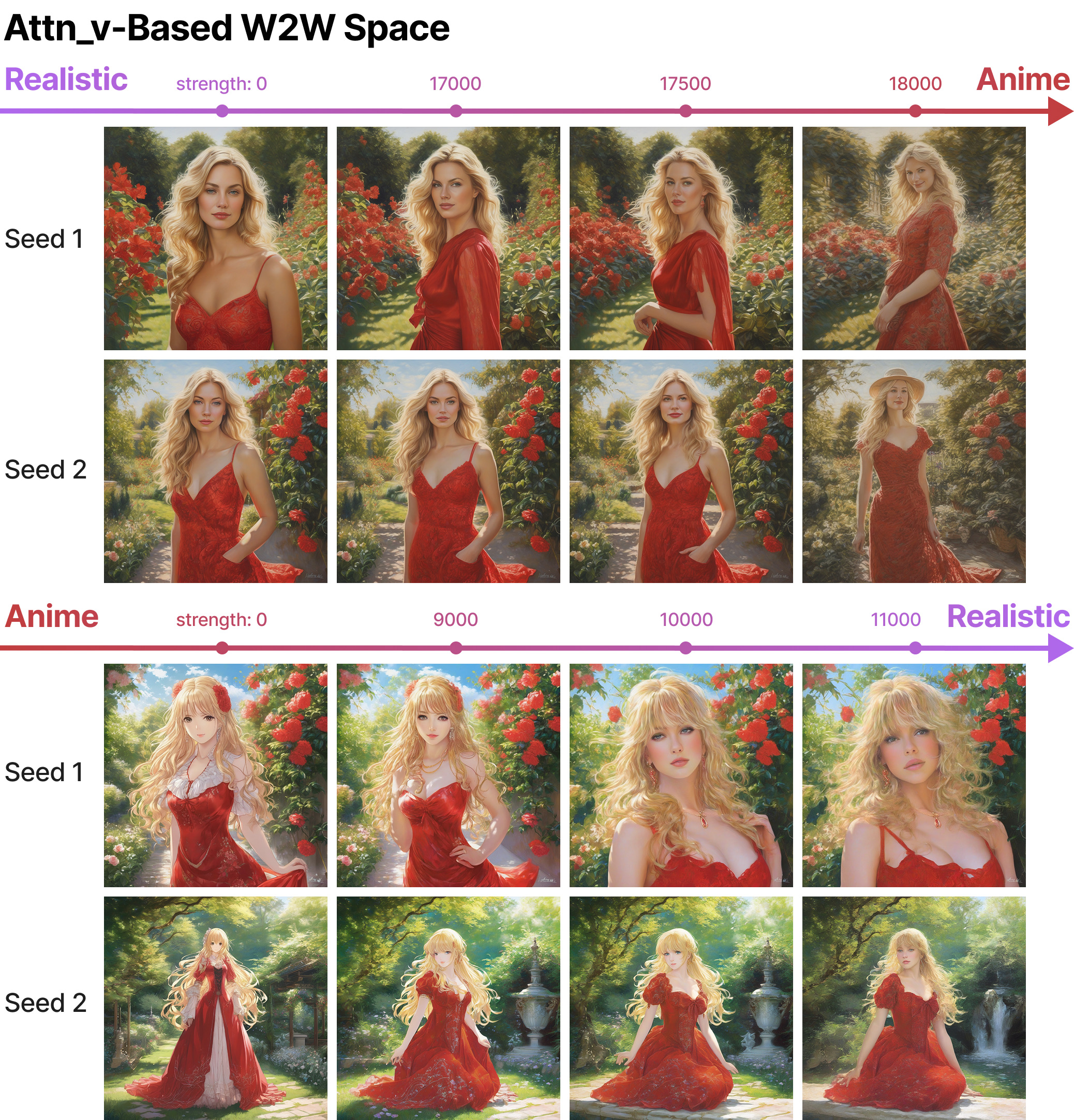}
    \caption{Editing results using the W2W space constructed from \texttt{attn\_v} layers. Top: transformation from \textit{realistic} to \textit{anime}. Bottom: transformation from \textit{anime} to \textit{realistic}. The first column shows outputs from the unedited base model; subsequent columns show results with increasing tuning strength. Each row shares the same generation seed. While the \textit{ani-real} direction produces coherent transitions, the reverse \textit{real-ani} direction is less effective.}
    \label{fig:attn_v_ani_real}
\end{figure*}

\begin{figure*}[ht]
    \centering
    \includegraphics[width=0.85\linewidth]{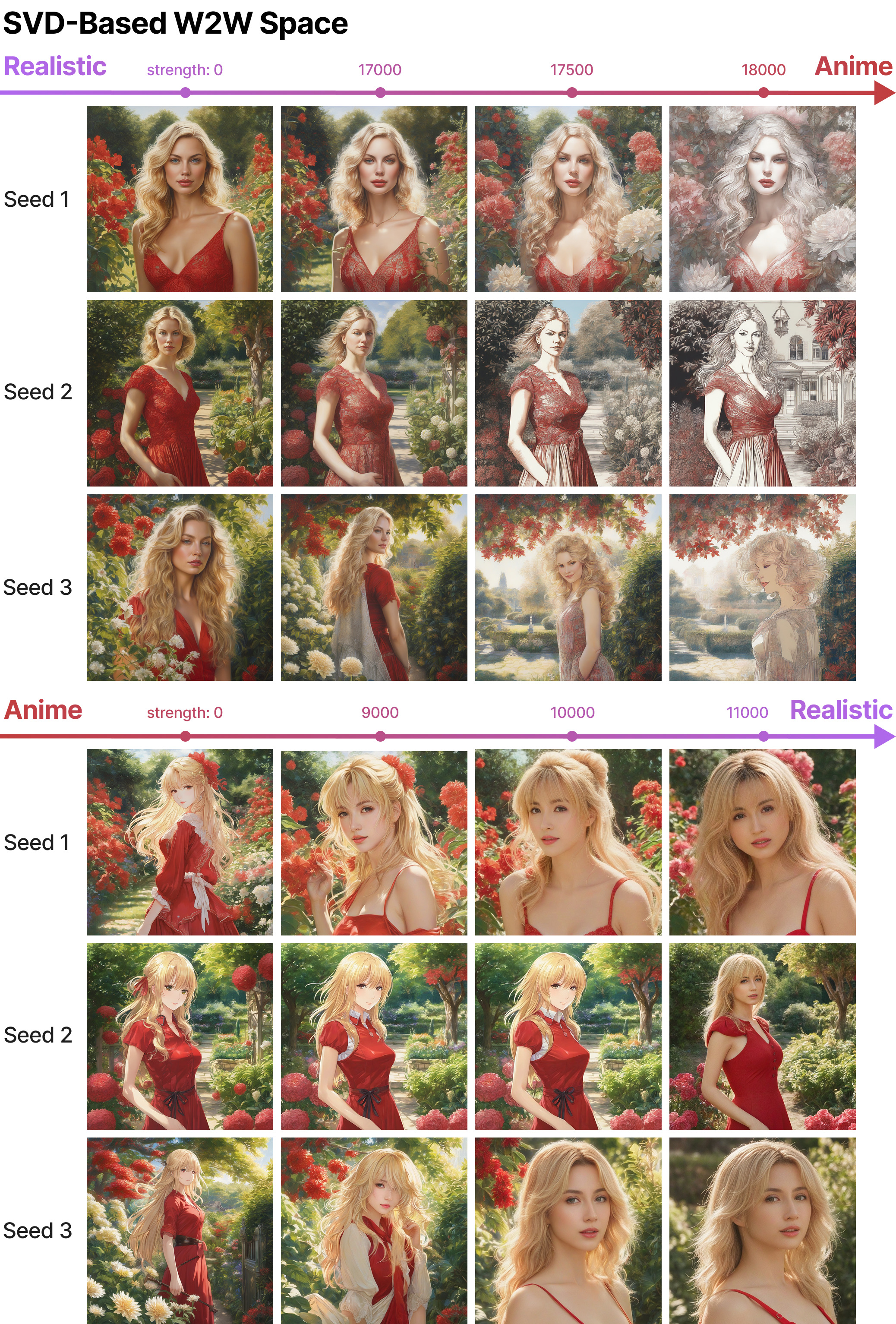}
    \caption{Editing results using the SVD-based W2W space. Top: transformation from \textit{realistic} to \textit{anime}. Bottom: transformation from \textit{anime} to \textit{realistic}. The base model outputs are shown in the first column, followed by results with increasing tuning strength. Each row uses a fixed generation seed. The SVD-based representation supports smooth, bidirectional editing with semantically coherent outputs in both directions.}
    \label{fig:svd_ani_real}
\end{figure*}

To evaluate the effectiveness of different W2W space construction strategies, we compare the performance of the SVD-based and attn\_v-based approaches on both the \textit{ani-real} and \textit{real-ani} directions. As shown in Fig.~\ref{fig:svd_ani_real}, the SVD-based W2W space enables smooth and coherent transformations in both directions. In contrast, the attn\_v-based W2W space performs well for \textit{ani-real} but fails to generalize to \textit{real-ani} (Fig.~\ref{fig:attn_v_ani_real}). These results underscore the superior bidirectional editing capability of the SVD-based approach.

\subsection{User Preference Description}
\begin{figure*}[ht]
    \centering
    \includegraphics[width=0.85\textwidth]{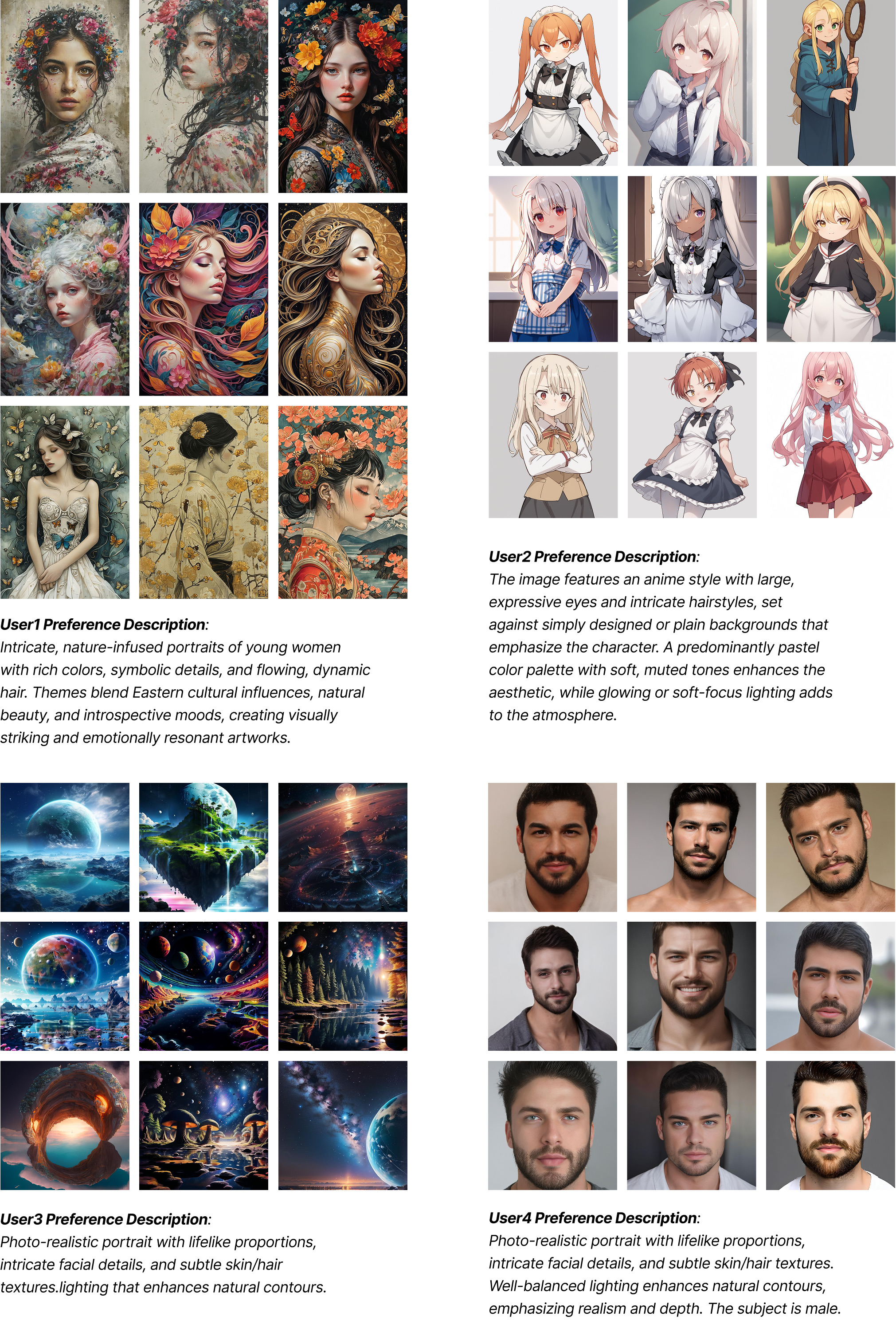}
    \caption{User TOP 9 preference images along with the textual descriptions}
    \label{fig:user_pref_desc}
\end{figure*}

Fig.\ref{fig:user_pref_desc} shows Top 9 preference images of user $P_1$,$P_2$,$P_3$,$P_4$, along with their corresponding textual descriptions.

\subsection{Multi-User Preference Alignment Results}
\label{sec:multi_user_appendix}
\begin{figure*}[ht]
    \centering
    \includegraphics[width=1\linewidth]{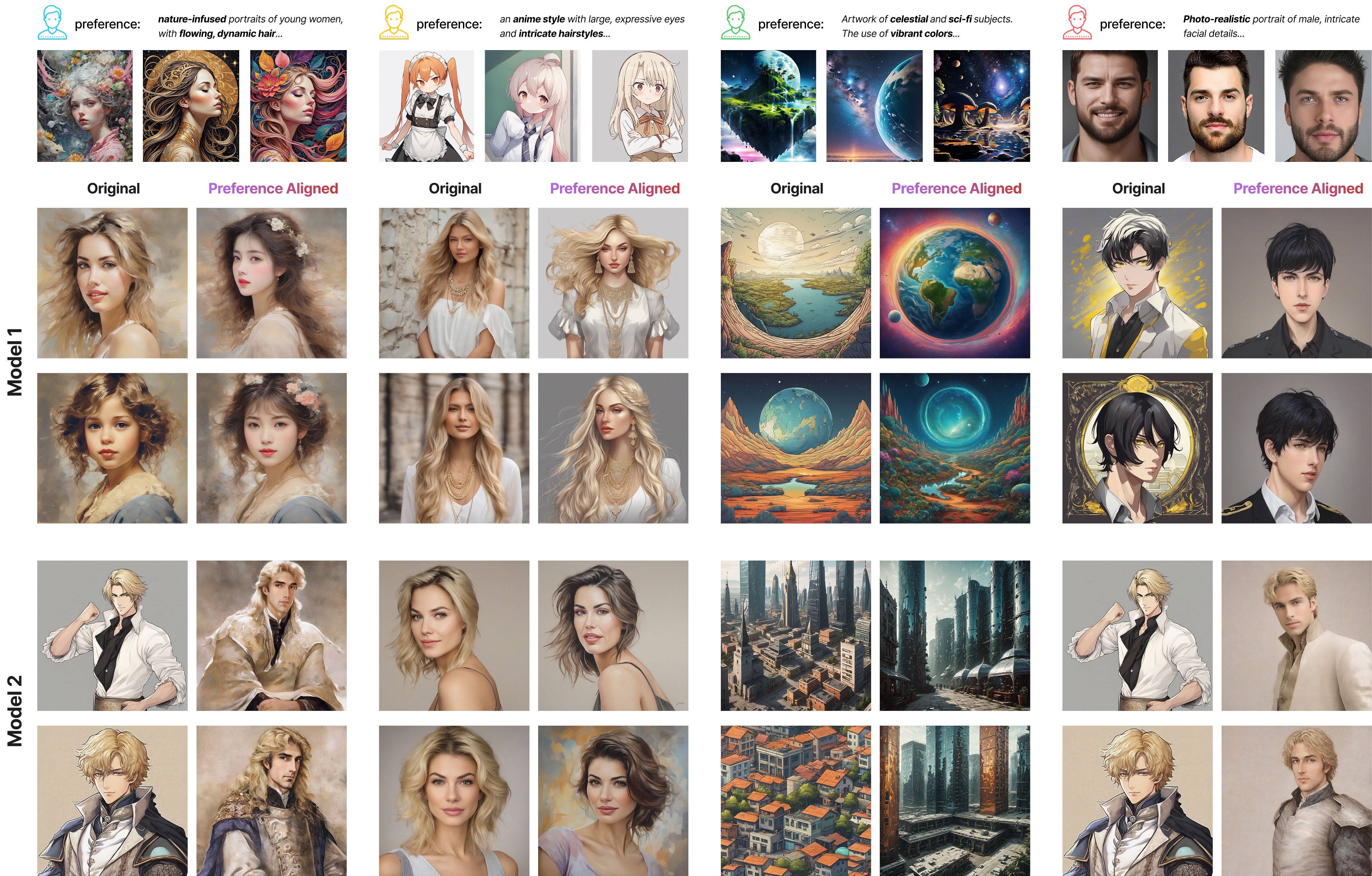}
    \caption{This figure illustrates the alignment of generative models to individual user preferences. Each user’s visual preference is shown at the top, with generated samples below. Left images are from the unedited SDXL base model; right images are from the edited models. 
    }
    \label{fig:multi_user}
\end{figure*}

Fig.\ref{fig:multi_user} demonstrates preference alignment for four users, where initial misaligned models were adjusted along learned directions. Beyond visual improvements, both the CLIP score and VLM-based rankings are higher for these edited images compared to the original outputs, confirming enhanced alignment after editing.

\subsection{Image Generation Implementation}
Tab.~\ref{tab:user_pref_generation_config} 
provides a comprehensive overview of the image generation settings for different users. It outlines the model versions used, specific prompts, seeds, and key parameters such as edit strength. All images for generative model personalization were generated as $1024\times 1024$px, with $30$ inference steps, guidance scale $5$, and LoRA scale $1$.

\newpage

\begin{table*}[h]
    \centering
    \scriptsize
    \setlength{\tabcolsep}{3pt}
    \renewcommand{\arraystretch}{1.2}
    \resizebox{\textwidth}{!}{
    \begin{tabular}{llp{6cm}cc}
        \toprule
        \textbf{User} & \textbf{Model Version ID} & \textbf{Prompt} & \textbf{Seeds} & \textbf{Edit Strength} \\
        \midrule
        \multirow{2}{*}{User1} & 315523 & portrait of a girl, high quality, ftsy-gld. & [2, 900] & 6000 \\
        & 150333 & a man, Prince Hamlet, blonde, cessa style, looking at viewers, half-body, simple background, simple outfit. & [2, 37480] & 7500 \\
        \midrule
        \multirow{2}{*}{User2} & 480560 & Dasha, with her blonde hair cascading over her shoulders and a delicate necklace accentuating her long hair. & [900, 7892] & 6500 \\
        & 802411 & portrait of a women, high quality, J4ck13RJ. & [2, 50] & 7500 \\
        \midrule
        \multirow{2}{*}{User3} & 179603 & view of planet earth from distant, cartooneffects one. & [2, 24] & 7000 \\
        & 565887 & view of some buildings, from a distant, high quality, detailed, secretlab. & [23, 37480] & 6000 \\
        \midrule
        \multirow{2}{*}{User4} & 577810 & portrait of a boy, high quality, linden de romanoff, black hair, yellow eyes, short hair, hair between eyes, bangs, simple background. & [10, 285891] & 6000 \\
        & 150333 & A man, Prince Hamlet, blonde, cessa style, looking at viewers, half-body, simple background, simple outfit. & [2, 3] & 6000 \\
        \bottomrule
    \end{tabular}
    }
    \caption{Generation settings for preference alignment}
    \label{tab:user_pref_generation_config}
\end{table*}

\cleardoublepage

\end{document}